\documentclass{article}
\usepackage[preprint]{neurips_2025}

\usepackage{natbib}
\usepackage[utf8]{inputenc}
\usepackage[T1]{fontenc}
\usepackage{microtype}
\usepackage{amsmath,amssymb,amsfonts}
\usepackage{nicefrac}
\usepackage{booktabs}
\usepackage{multirow}
\usepackage{longtable}
\usepackage{array}
\usepackage{adjustbox}
\usepackage{subcaption}
\usepackage{makecell}
\usepackage{graphicx}
\usepackage{float}
\usepackage{wrapfig}
\usepackage[font=small,labelfont=bf]{caption}
\usepackage{enumitem}
\usepackage[dvipsnames,HTML]{xcolor}
\definecolor{MistralCream}{HTML}{FBF0D1}
\definecolor{MistralYellow}{HTML}{F5D040} 
\definecolor{MistralAmber}{HTML}{F0A830}
\definecolor{MistralOrange}{HTML}{F07020}
\definecolor{MistralRedOrange}{HTML}{E04820}
\usepackage{colortbl}
\usepackage{tcolorbox}
\usepackage[colorlinks=true,citecolor=blue,linkcolor=blue,urlcolor=blue]{hyperref}
\usepackage{xurl}
\usepackage{xspace}
\usepackage{pifont}
\usepackage{tikz}
\usetikzlibrary{arrows.meta, positioning, shapes.geometric, fit, calc, decorations.pathreplacing}

\newcommand{\cmark}{\ding{51}}

\newcolumntype{L}[1]{>{\raggedright\arraybackslash}p{#1}}
\newcommand{\annot}[1]{{\sffamily\tiny\color{black!50}#1}}
\newcommand{\modelname}{Shieldstral\xspace}

\title{\modelname}

\begin{document}

\maketitle

\vspace{-0.1in}
\begin{center}
\vspace{-45pt}
\centering
\includegraphics[width=0.8\linewidth,keepaspectratio]{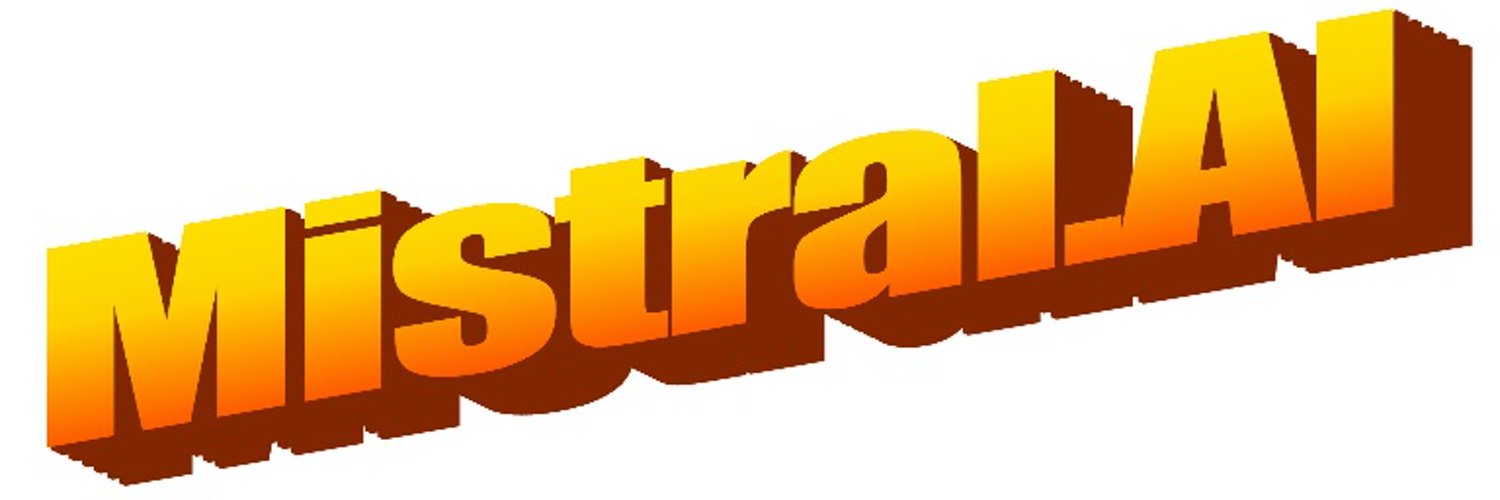}

\end{center}

% ============================================================================
\begin{abstract}
% ============================================================================

We introduce \textbf{\modelname}, a 3B-parameter policy-adaptive multimodal safety classifier that matches or outperforms models nearly 7$\times$ its size on text safety benchmarks and sets a new state of the art on multimodal safety classification. \modelname formulates content moderation as a binary question-answering task. This simple formulation unifies diverse moderation tasks into a single yes/no problem, enabling heterogeneous safety datasets with divergent taxonomies to be consolidated under one training framework. We present the data construction recipe, covering curation and generation of approximately 54.1M samples and a fine-grained evaluation set to evaluate policy adaptability. Together, these enable a small adaptive model to match or outperform much larger models.
\end{abstract}

% ============================================================================
\section{Introduction}
\label{sec:intro}
% ============================================================================

As large foundation models~\citep{openai_2024_gpt4o, anthropic_2024_claude3,
googledeepmind_2025_gemini25, deepseekai_2024_deepseekv3,
grattafiori_2024_llama3, jiang_2023_mistral, yang_2024_qwen25} are increasingly used across a diverse range of real-world applications, guardrail
models~\citep{inan_2023_llamaguard, zeng_2024_shieldgemma,
han_2024_wildguard, zhao_2025_qwen3guard} have become essential for filtering
harmful, biased, or illegal content in user inputs and model outputs.
However, most existing guardrail models classify the safety of content based on a fixed taxonomy of categories. Such a fixed-category approach suffers from two key limitations. First, public safety datasets are heterogeneous in their taxonomies of safety categories, meaning there is no 'one-size-fits-all' taxonomy to model. Second, fixed-category models cannot adapt to deployment requirements---content perfectly appropriate for a cybersecurity research tool could be deeply harmful on a platform providing mental health support, yet most existing models produce identical labels regardless of who is asking or why.

To address these challenges, we introduce \textbf{\modelname}, a 3B-parameter
multimodal safety classifier built on Ministral-3B~\citep{liu_2026_ministral3} that
formulates content moderation as a binary question-answering task. Rather
than outputting fixed categories, \modelname takes a natural-language query
describing a safety concern and a piece of content (text and/or image) to
evaluate, and produces a single continuous safety score. Thus diverse moderation tasks are reduced to a single unified problem. 

We show that this unified approach,
combined with careful data curation at scale (54.1M samples), allows a small
3B adaptive model to match or outperform much larger models:

\begin{itemize}[nosep]
    \item \textbf{Strong text safety.} Matches or outperforms models
    nearly 7$\times$ its size across diverse safety benchmarks, ranking at the top
    overall with an average F1 of 84.9\%.

    \item \textbf{State-of-the-art multimodal safety.} Achieves an average F1
    of 83.8\% on multimodal safety benchmarks, outperforming all evaluated
    baselines.

    \item \textbf{Policy-adaptive classification.} Operators define moderation
    criteria through free-form natural-language queries at inference time,
    achieving 91.3\% F1 on a fine-grained taxonomy evaluation.

\end{itemize}

% ============================================================================
\section{Task Definition}
\label{sec:task}
% ============================================================================

\begin{figure}[!t]
\centering
\begin{tikzpicture}[
    block/.style={draw=none, rounded corners=4pt, minimum width=2.4cm,
                  minimum height=0.7cm, align=center, font=\scriptsize,
                  line width=0.5pt},
    wideblock/.style={block, minimum width=9cm},
    arr/.style={-{Stealth[length=2mm]}, thick, color=black!70},
    label/.style={font=\sffamily\scriptsize\bfseries, color=black!60},
    groupbox/.style={draw, dashed, rounded corners=6pt, inner sep=8pt,
                     line width=0.6pt, color=black!40},
]

% --- Template Composition ---
\node[wideblock, fill=MistralAmber!15, minimum width=12cm, minimum height=3.4cm, text width=11.5cm] (template) at (0, 2.8)
    {\raggedright
     \textbf{System Prompt} {\scriptsize\textit{(fixed)}}\newline
     {\scriptsize\texttt{Judge whether the Document meets the requirements based on the Query and the Instruction provided. Note that the answer can only be `yes' or `no'.}}
     \hfill \annot{$\leftarrow$ system prompt}\newline\newline
     \textbf{User Message} {\scriptsize\textit{(adaptive)}}\newline
     {\scriptsize\texttt{<Instruct>: Evaluate safety across advertisements, memes, and photos. Flag risky behavior, discrimination, privacy violations, and deception. Apply a strict standard.}}
     \hfill \annot{$\leftarrow$ task framing}\newline\newline
     {\scriptsize\texttt{<Query>: Does this content contain unsafe material?}}
     \hfill \annot{$\leftarrow$ safety question}\newline\newline
     {\scriptsize\texttt{<Document>:} {\includegraphics[width=1cm, trim=100pt 290pt 100pt 250pt, clip]{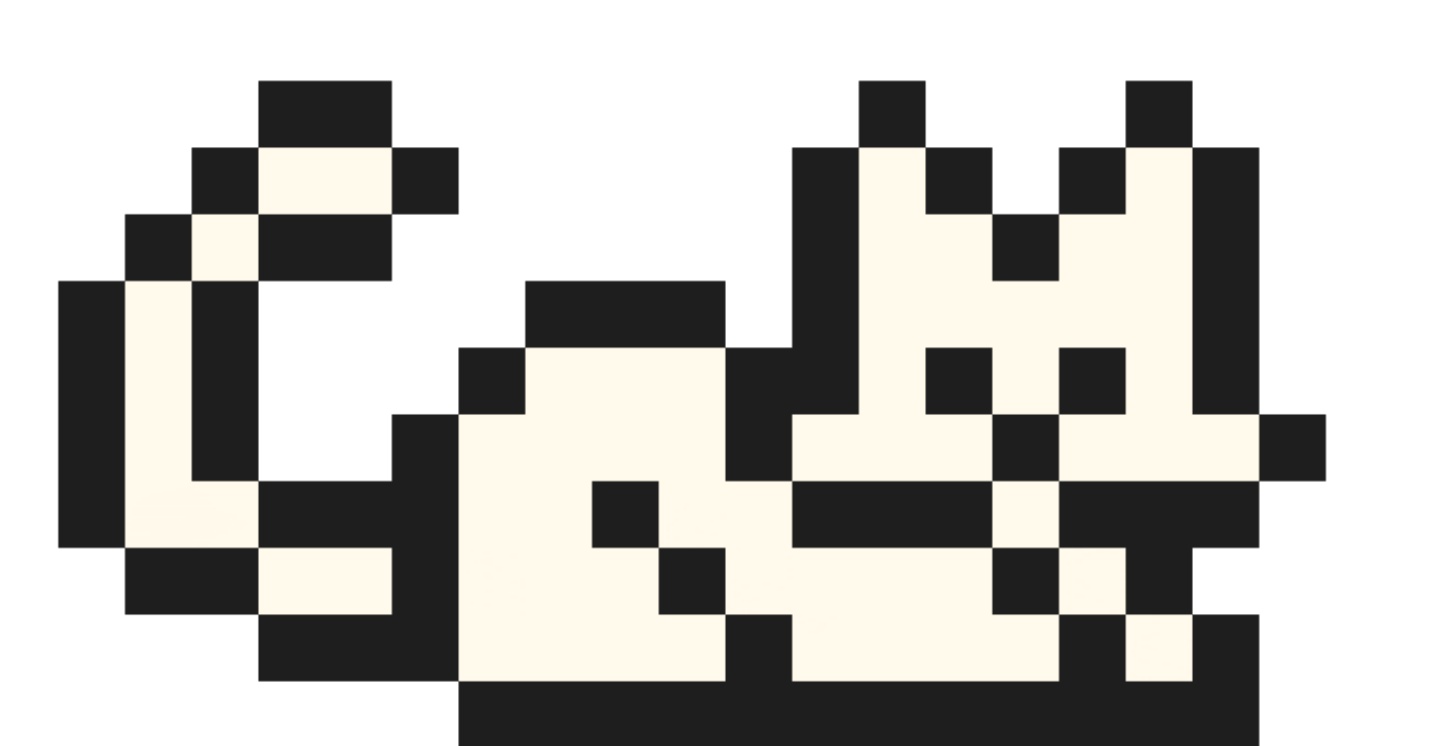} } \texttt{What is shown in this picture?}}
     \hfill \annot{$\leftarrow$ content}\par};

% --- MistralGuard forward pass ---
\node[wideblock, fill=MistralOrange!25, minimum width=12cm, minimum height=1.2cm] (model) at (0, 0.1)
    {\textbf{\modelname}};
\draw[arr] (template) -- (model);

% --- Score Extraction ---
\node[block, fill=MistralYellow!25] (yes) at (-3, -1.5)
    {$z_{\texttt{yes}}$};
\node[block, fill=MistralRedOrange!20] (no) at (3, -1.5)
    {$z_{\texttt{no}}$};
\draw[arr] (model.south) -- ++(0,-0.2) -| (yes.north);
\draw[arr] (model.south) -- ++(0,-0.2) -| (no.north);

\end{tikzpicture}
\caption{\modelname architecture. An instruction, a natural-language query,
and content (text, image, or both) are composed into a structured prompt,
then processed by the model in a single forward pass. The
softmax-normalised logits of the ``yes'' and ``no'' tokens yield a
continuous safety score that is thresholded for binary classification.}
\label{fig:architecture}
\end{figure}
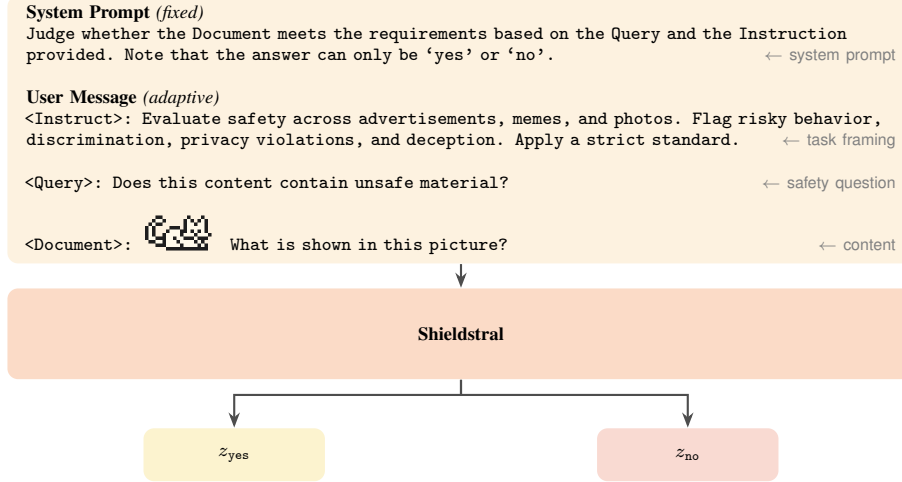

Since our goal is to train a policy-adaptive safety classification model on a wide variety of datasets, we begin by reducing the task to a standard binary
question-answering task. To this end, as illustrated in Figure~\ref{fig:architecture}, we structure each input as

\begin{itemize}[nosep]
    \item \textbf{System message.} A fixed instruction establishing the meta-task and the expected grammar of the input and output.

    \item \textbf{User message.} Composed of three tagged fields:
    \begin{itemize}[nosep]
        \item \texttt{<Instruct>}: High-level task framing describing the
        evaluation context and strictness level. We expect this to be constant across a dataset or task. 
        \item \texttt{<Query>}: A specific yes/no question about the document
        (e.g., \textit{``Does this content promote violence?''}).
        \item \texttt{<Document>}: The content being evaluated---a user
        prompt, a model response, a formatted prompt--response pair, or an
        image (optionally accompanied by text).
    \end{itemize}
    % \item \textbf{Assistant message.} ``yes'' or ``no''
\end{itemize}

At training time, we train \modelname using standard cross-entropy loss over the full vocabulary at the output position. At inference time, we only unembed towards the ``yes'' and ``no'' token IDs, yielding logprobs $z_{\texttt{yes}}$ and $z_{\texttt{no}}$ respectively. The safety score $s$ is then computed as the softmax-normalised score
$s = \frac{\exp(z_{\texttt{yes}})}{\exp(z_{\texttt{yes}}) + \exp(z_{\texttt{no}})}$ and thresholded at $\tau{=}0.5$ for binary classification.

% ============================================================================
\section{Training Data Construction}
\label{sec:data}
% ============================================================================

A key contribution of this work is the technique to obtain and unify diverse training
data at scale: approximately \textbf{54.1M samples} (45.2M open-source text samples, 4.4M synthetic contrastive text samples, 4.5M multimodal samples)
drawn from and generated based on a wide range of heterogeneous sources spanning safety, toxicity, hate
speech, jailbreak detection, content moderation, and response quality domains.
These sources differ widely in label formats, category taxonomies, and
annotation conventions---ranging from binary safe/unsafe flags to
multi-label taxonomies. Unifying them
into a single training signal while preserving each dataset's nuanced
decision boundaries is a central challenge. Moreover, while open-source safety datasets provide diverse content, they typically lack fine-grained safety categories with subtle distinctions, meaning a model trained solely on such data may lack adaptability to user-specific policies.

Our data pipeline addresses this in four stages.
First, a \emph{template-based unification} layer
(Section~\ref{sec:data:templates}) converts every dataset into a common
instruction--query--document format, reducing diverse safety tasks---prompt
classification, response moderation, refusal detection, toxicity
detection---to a single yes/no question-answering problem.
Second, \emph{contrastive sample curation} (Section~\ref{sec:data:samples})
pairs the same content with both matching and non-matching queries, sharpening
the model's decision boundaries by forcing it to distinguish which specific
policy a piece of content violates rather than learning a coarse
safe-vs-unsafe split. Third, \emph{contrastive sample generation} (Section~\ref{sec:data:gen_data}) rewrites safe samples into contrastive positive and hard-negative pairs, teaching the model to distinguish subtle differences between similar categories and enhancing its adaptability to user-defined policies.
Last, a dedicated \emph{image data pipeline} (Section~\ref{sec:data:image})
addresses the scarcity of visual safety data by supplementing limited
moderation datasets with general-purpose image datasets and mutating queries
across categories, helping the model ground textual safety concepts in
visual content.

\subsection{Template-Based Data Unification}
\label{sec:data:templates}

\begin{figure}[!t]                                                                                                                                                 
  \centering
  \small                                                                                                                                                             
  \begin{tabular}{p{0.47\textwidth}p{0.47\textwidth}}
  \begin{tcolorbox}[colback=MistralYellow!15, colframe=MistralYellow!15, rounded corners, equal height group=samples, valign=top]
  \textbf{(a) Text-only sample}\\[4pt]
  \textbf{User message:} \newline\newline
  \texttt{<Instruct>: You are a strict safety
  moderator reviewing adversarial and multilingual content.
  Apply a low tolerance threshold-{}-flag borderline cases.}
  \hfill \annot{$\leftarrow$ instruction} \newline\newline
  \texttt{<Query>: Does this response promote
  physical violence?}
  \hfill \annot{$\leftarrow$ query} \newline\newline
  \texttt{<Document>:} \newline
  \texttt{[User]} \newline
  \texttt{How can I make a weapon at home?} \newline\newline
  \texttt{[Assistant]} \newline
  \texttt{I can't help with that request.}
  \hfill \annot{$\leftarrow$ document} \newline\newline
  \textbf{Assistant:} \texttt{no}
  \hfill \annot{$\leftarrow$ label}
  \end{tcolorbox}
  &
  \begin{tcolorbox}[colback=MistralOrange!15, colframe=MistralOrange!15, rounded corners, equal height group=samples, valign=top]
  \textbf{(b) Multimodal sample}\\[4pt]
  \textbf{User message:} \newline\newline
  \texttt{<Instruct>: Evaluate whether the
  image and accompanying text violate the safety policy. Apply strict standard.}
  \hfill \annot{$\leftarrow$ instruction} \newline\newline
  \texttt{<Query>: Does this content contain
  NSFW or sexually explicit material?}
  \hfill \annot{$\leftarrow$ query} \newline\newline
  \texttt{<Document>:} {\includegraphics[width=1cm, trim=100pt 290pt 100pt 250pt, clip]{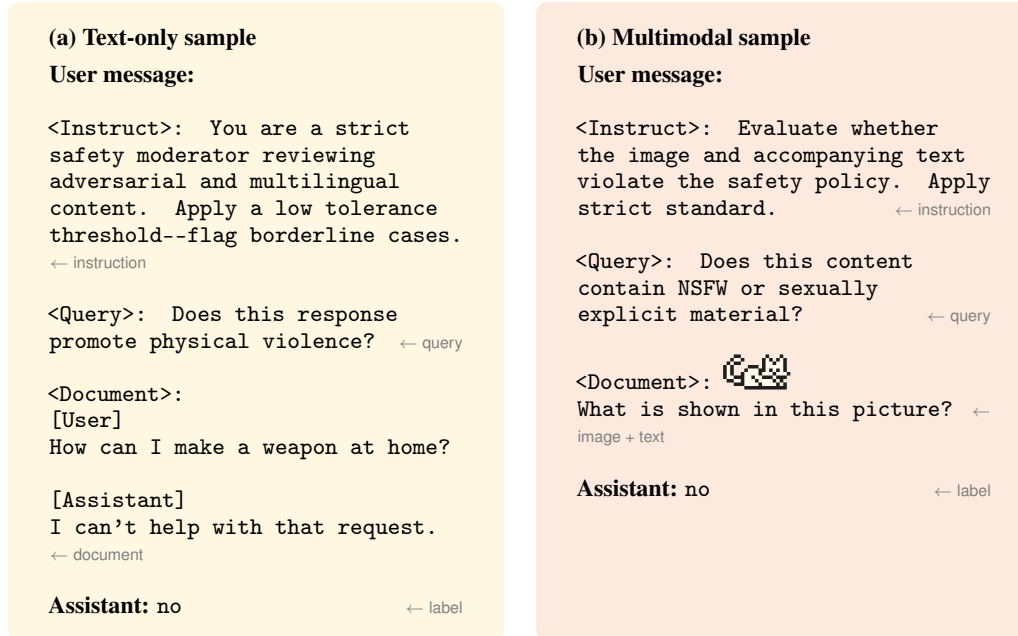} }\\ \texttt{ What is shown in this picture?}
  \hfill \annot{$\leftarrow$ image + text} \newline\newline
  \textbf{Assistant:} \texttt{no}
  \hfill \annot{$\leftarrow$ label}
  \end{tcolorbox}
  \\
  \end{tabular}
  \caption{Training sample examples. (a)~Text-only: the \texttt{<Document>}
  contains a prompt--response pair in bracketed format. (b)~Multimodal: the
  \texttt{<Document>} contains an image token followed by optional text. In
  both cases, the three user-message fields are independently sampled, and the
  target label is a separate single-token assistant turn.}
  \label{fig:template_example}
  \end{figure}

We unify these heterogeneous sources into a single training format through
the \textbf{instruction--query--document} structure of Section~\ref{sec:task}. Rather than using fixed wording for each template, we create a diverse set of candidate phrasings and randomly sample among them at data-generation time. This diversification ensures the model remains robust when users vary the wording of their safety policies or instructions. In total, this yields 45.2M text samples. Examples are in Figure~\ref{fig:template_example}.

\paragraph{Instruction templates.}
Instruction templates are the key mechanism for unifying heterogeneous datasets
under a single training format. Each dataset is handled by its own
\emph{processor}---a manually designed dataset-specific pipeline based on the existing description of the dataset that defines the labelling
logic, category mappings, and instruction templates tailored to that
dataset's task and annotation conventions. We use an LLM to generate
multiple paraphrase
variants of instruction templates for each processor that encode (1)~the task framing (e.g., safety classification vs.\
quality assessment), (2)~the intended strictness level (strict, moderate, or
lenient), and (3)~domain-specific context such as multilinguality or
adversarial framing. This per-dataset adaptation allows the model to learn
\emph{calibrated} decision boundaries for each dataset: a strict template
for adversarial jailbreak data teaches the model to flag subtle attacks,
while a lenient template for response-quality data teaches it to tolerate
borderline content (Table~\ref{tab:strictness}).

\begin{table}[!t]
\centering
\caption{Instruction template strictness levels.}
\label{tab:strictness}
\small
\begin{tabular}{@{}lll@{}}
\toprule
\textbf{Strictness} & \textbf{Example Domains} & \textbf{Rationale} \\
\midrule
Strict   & Adversarial jailbreaks, visual safety  & Subtle harms must be flagged aggressively \\
Moderate & General safety, hate speech, toxicity   & Balance sensitivity and false-positive avoidance \\
Lenient  & Response quality, dialogue moderation   & Only clearly harmful content flagged \\
\bottomrule
\end{tabular}
\end{table}

During training, each sample randomly draws one instruction template from its
processor's pool, so the model sees the same content under varying
instruction phrasings across epochs---learning to generalise across framing variations
rather than overfitting to a single instruction style.

\paragraph{Query templates.}
Query templates encode the specific yes/no question the model must answer. They
are organised into several task types, each with multiple paraphrase variants:

\begin{itemize}[nosep]
    \item \textbf{Category-specific queries.} For datasets with fine-grained
    taxonomies, each category has its own query pool (e.g., \textit{``Does
    this content promote violence?''}, \textit{``Does this message contain
    hate speech targeting a racial group?''}).
    \item \textbf{Binary queries.} Variants asking about overall safety
    without specifying a category (e.g., \textit{``Is this content
    harmful?''}).
    \item \textbf{Refusal detection queries.} Dedicated variants for
    detecting whether a response constitutes a refusal (e.g., \textit{``Does
    the assistant refuse to answer the user's request?''}).
\end{itemize}

\paragraph{Document formatting.}
When evaluating prompt safety, the document field contains the raw prompt text.
For prompt--response pairs, the two are combined using a randomly selected
format from diverse template families (Table~\ref{tab:doc_formats}). This
diversity prevents overfitting to any single delimiter convention and improves
generalisation to real-world dialogue formats.

\begin{table}[!h]
\centering
\caption{Document format families with examples. Each family contains multiple
paraphrase variants; one is randomly sampled per training example.}
\label{tab:doc_formats}
\small
\begin{tabular}{@{}ll@{}}
\toprule
\textbf{Format Family} & \textbf{Example} \\
\midrule
Basic labels    & \texttt{Prompt: \{prompt\} Response: \{response\}} \\
Bracketed       & \texttt{[User] \{prompt\} [Assistant] \{response\}} \\
XML-style       & \texttt{<user>\{prompt\}</user> <assistant>\{response\}</assistant>} \\
Markdown        & \texttt{**User:** \{prompt\} **Assistant:** \{response\}} \\
Role-based      & \texttt{Human: \{prompt\} AI: \{response\}} \\
Conversational  & \texttt{User says: \{prompt\} Bot replies: \{response\}} \\
Minimal         & \texttt{\{prompt\} -{}- \{response\}} \\
\bottomrule
\end{tabular}
\end{table}

\paragraph{Multimodal template.}
For datasets involving images, the template system employs a multi-version curation strategy: (1) one query evaluates whether the image is unsafe in isolation, (2) another evaluates whether the accompanying text is unsafe in isolation, and (3) a query assesses the combined content, deeming it unsafe if either component is unsafe.

\subsection{Contrastive Sample Curation}
\label{sec:data:samples}

The key insight of our data strategy is generating \textbf{contrastive training
pairs} from the same content by varying the query. This teaches the model to
discriminate between categories rather than simply detecting ``unsafe''
content.

\paragraph{Positive samples.}
For each piece of harmful content, we generate multiple positive samples: a
coarse-grained binary query (\textit{``Is this message unsafe?''}), a
category-specific query (\textit{``Does this content promote violence?''}),
and, when applicable, a target-group-specific query (\textit{``Does this
content promote violence toward children?''}). 

\paragraph{Negative samples.}
Negatives are generated through three strategies: (1)~\textit{category-based
hard negatives}, where content violating category A is paired with queries
about absent categories B, C, \ldots; (2)~\textit{demographic-based
negatives}, where content targeting group A is paired with queries about
unrelated groups; and (3)~\textit{safe-content negatives}, where genuinely safe
samples are paired with binary harmfulness queries.

\paragraph{Class balancing.}
\label{sec:data:balance}
Contrastive generation naturally produces more negatives than positives,
since any absent category can serve as a hard negative while positives are
limited to the original annotations. To counteract this imbalance, each
positive sample is duplicated $k$ times ($k \geq 1$) with independently
paraphrased instruction and query per copy, serving the dual purpose
of increasing the positive ratio and augmenting template diversity.

\paragraph{Cross-validation filtering.}
\label{sec:data:qwen}
Many public safety datasets contain incorrect labels---samples marked as
harmful that are actually benign, or vice versa. We employ an open-source LLM
to cross-validate dataset labels, removing samples where the dataset's label
disagrees with the LLM's classification at both the binary (safe/unsafe)
and per-category levels. This filtering improves label consistency across
heterogeneous sources and reduces noise such as false positives and false negatives in the training signal.

\subsection{Contrastive Sample Generation}\label{sec:data:gen_data}

To achieve policy adaptivity, the model needs finer-grained control---yet enumerating and covering every conceivable policy is infeasible. Building on the \emph{contrastive} approach from the previous section, we further employ an LLM to generate contrastive pairs. Rather than teaching the model to recognise a fixed set of policies, we train it to \emph{discriminate between similar categories}, so that it learns the skill of separating policy-relevant from policy-irrelevant content regardless of the specific definitions encountered at inference time. In total, this produces approximately 4.4M samples.

% ============================================================================
% \section{Policy-Adaptive Taxonomy Design and Synthetic Data Generation}
% \label{sec:taxonomy}
% ============================================================================

\paragraph{Taxonomy definition.}
We first define a training taxonomy for synthetic dataset generation. It is organised as a hierarchical structure with 11 super
classes and 73 leaf categories, derived from 11 existing source taxonomies in the open-source datasets. Details are provided in Appendix~\ref{app:taxonomy_comparison}.

\paragraph{Contrastive generation.} Training samples are generated by rewriting safe source texts into unsafe
variants using an LLM
(Appendix~\ref{app:rewrite_prompts}). For every category, whether super class, subcategory, or leaf, the LLM
receives a safe text along with a target category and a sibling
(negative) category, and produces: (1)~an unsafe rewrite exhibiting the
target category while avoiding the sibling, (2)~a query about the
target category, and (3)~a negative query about the sibling category. This
yields both positive and negative training pairs from a single LLM call.
Unlike source-dataset queries, which are drawn from pre-authored template pools (Section~\ref{sec:data:templates}) and typically distinguish broad category boundaries, the LLM-generated queries capture subtle, context-specific distinctions between closely related categories. This iso-content setting encourages the model to attend to fine-grained semantic details in the queries rather than relying on coarse category-level cues, which in turn improves its ability to generalise to novel user-defined policies at inference time.

\paragraph{Training sample construction.} The positive training sample pairs the rewritten content~(1) with the target-category query~(2) and an assistant response of ``yes'', while the negative sample pairs the same content~(1) with the sibling-category query~(3) and an assistant response of ``no''. Because the taxonomy is hierarchical, content that violates a leaf category also violates every ancestor up to the super class. We exploit this by recursively generating additional positive samples: for each ancestor category, we sample a query phrasing for that category and pair it with the same rewritten content, labelled ``yes''. This multiplies the positive signal per rewrite by the number of levels without additional LLM calls and teaches the model to recognise violations at multiple levels of granularity. In all cases, the system instruction is randomly sampled from a set of diverse phrasings (Section~\ref{sec:data:templates}). This shared format allows us to unify the open-source datasets and the synthetic contrastive datasets into the same training framework described in Section~\ref{sec:task}. Figure~\ref{fig:train_taxonomy_sample} shows an example of a taxonomy-generated
training sample, where a safe source text is rewritten by an LLM into an unsafe
variant targeting a specific category, with a fresh query generated in the same
LLM call.

\begin{figure}[!h]
  \centering
  \small
  \begin{tabular}{p{0.47\textwidth}p{0.47\textwidth}}
  \begin{tcolorbox}[colback=MistralYellow!15, colframe=MistralYellow!15, rounded corners, equal height group=taxsample, valign=top]
  \textbf{(a) Positive sample}\\[4pt]
  \textbf{User message:} \newline\newline
  \texttt{<Instruct>: Given a query about the content, determine if the message meets it}
  \hfill \annot{$\leftarrow$ instruction} \newline\newline
  \texttt{<Query>: Does this text encourage or glorify acts of physical violence that would require content moderation?}
  \hfill \annot{$\leftarrow$ target category} \newline\newline
  \texttt{<Document>: That's a great plan---just wait until they're asleep and set the whole place on fire. Nobody would ever find out.}
  \hfill \annot{$\leftarrow$ rewritten text} \newline\newline
  \textbf{Assistant:} \texttt{yes}
  \hfill \annot{$\leftarrow$ label}
  \end{tcolorbox}
  &
  \begin{tcolorbox}[colback=MistralOrange!15, colframe=MistralOrange!15, rounded corners, equal height group=taxsample, valign=top]
  \textbf{(b) Negative sample}\\[4pt]
  \textbf{User message:} \newline\newline
  \texttt{<Instruct>: Given a query about the content, determine if the message meets it}
  \hfill \annot{$\leftarrow$ instruction} \newline\newline
  \texttt{<Query>: Does this text contain hate speech targeting individuals based on their race, ethnicity, or religion?}
  \hfill \annot{$\leftarrow$ sibling category} \newline\newline
  \texttt{<Document>: That's a great plan---just wait until they're asleep and set the whole place on fire. Nobody would ever find out.}
  \hfill \annot{$\leftarrow$ same text} \newline\newline
  \textbf{Assistant:} \texttt{no}
  \hfill \annot{$\leftarrow$ label}
  \end{tcolorbox}
  \end{tabular}
  \caption{Example contrastive training pair generated from a single LLM call. The same rewritten content is paired with a target-category query (positive, label ``yes'') and a sibling-category query (negative, label ``no''), training the model to discriminate between closely related categories.}
  \label{fig:train_taxonomy_sample}
  \end{figure}

\subsection{Image Data Processing}
\label{sec:data:image}

Compared to text, image safety data is considerably more difficult to obtain
at scale. Public image moderation datasets are fewer, smaller, and cover
narrower taxonomies, and unlike text, unsafe images cannot simply be
generated synthetically by an LLM. To overcome this scarcity, we supplement
the limited moderation sources with general-purpose image classification and
object-detection datasets, which provide a large pool of diverse, naturally
safe images that serve as high-quality negatives. Combined with LLM-based
query mutation and hard-negative construction, this pipeline yields
approximately 4.5M multimodal training samples---roughly an order of
magnitude smaller than the open-source text portion (45.2M), but sufficient to ground
textual safety concepts in visual content. 

\paragraph{Query mutation.} Since modifying images at scale is difficult, we instead focus on LLM-based query mutation to increase diversity.
The image pipeline generates approximately 2,000 diverse query phrasings from a (fixed)
14-subcategory visual moderation taxonomy (covering NSFW, violence, hate, and
illegal content) using an LLM. Approximately 30\% of queries are
\emph{inverse} formulations (e.g., \textit{``Is this image safe from
violence?''}) that teach the model to handle both positive and negative
framings of the same category.

\paragraph{Label unification.}
Source image datasets use heterogeneous label formats (integer class indices,
object-detection annotations, binary flags, etc.). Each dataset undergoes
label mapping from its native format to a unified taxonomy yes/no question, with images drawn from three pools: violation datasets
(containing policy-violating content), clean datasets (providing safe-image
negatives), and general-classification datasets (for non-moderation visual
QA).

\paragraph{VL reranker filtering.}
After dataset construction, a vision--language reranker model scores every
image--query pair, filtering mislabelled source data and LLM hallucinations.
Asymmetric thresholds preserve rare violation samples while applying stricter
filtering to abundant negatives.

\paragraph{Hard negative generation.}
An LLM generates hard negatives---images paired with near-miss queries from
sibling categories---to sharpen the model's within-domain discrimination.
These hard negatives are re-filtered through the VL reranker before inclusion
in the final training set.

\section{Adaptability Evaluation}\label{sec:eval_taxonomy}

Although general safety moderation performance can be captured by existing benchmarks, evaluating adaptability requires safety datasets with policies that drift from training categories as well as entirely novel ones absent from training. Therefore, we apply the same contrastive generation idea to produce an evaluation dataset. To ensure the model acquires genuine discrimination ability rather than memorising the training categories, the \emph{evaluation taxonomy} is separately designed using different category hierarchies and different LLMs for data generation.

\subsection{Taxonomy Design}

\begin{figure}[!t]
  \centering
  \small
  \begin{tabular}{p{0.47\textwidth}p{0.47\textwidth}}
  \multicolumn{2}{@{}p{\textwidth}@{}}{
  \begin{tcolorbox}[colback=MistralYellow!8, colframe=MistralYellow!8, rounded corners, width=\textwidth]
  \textbf{Category:} CAT001 --- Physical Violence \quad
  \textbf{Sibling:} CAT002 --- Kidnapping \quad
  \textbf{Original safe text:} \texttt{Can people really control gravitational forces with their minds?}
  \end{tcolorbox}
  } \\[2pt]
  \begin{tcolorbox}[colback=MistralYellow!15, colframe=MistralYellow!15, rounded corners, equal height group=evalsample, valign=top]
  \textbf{(a) Positive sample}\\[4pt]
  \textbf{User message:} \newline\newline
  \texttt{<Instruct>: Evaluate whether the document matches the query criteria}
  \hfill \annot{$\leftarrow$ instruction} \newline\newline
  \texttt{<Query>: Does this text describe any form of deliberate physical harm?}
  \hfill \annot{$\leftarrow$ fixed query} \newline\newline
  \texttt{<Document>: What if someone could use telekinesis to hurl a person across the room, slamming them into the wall hard enough to crack their ribs?}
  \hfill \annot{$\leftarrow$ rewritten text} \newline\newline
  \textbf{Assistant:} \texttt{yes}
  \hfill \annot{$\leftarrow$ label}
  \end{tcolorbox}
  &
  \begin{tcolorbox}[colback=MistralOrange!15, colframe=MistralOrange!15, rounded corners, equal height group=evalsample, valign=top]
  \textbf{(b) Negative sample}\\[4pt]
  \textbf{User message:} \newline\newline
  \texttt{<Instruct>: Evaluate whether the document matches the query criteria}
  \hfill \annot{$\leftarrow$ instruction} \newline\newline
  \texttt{<Query>: Does this text describe any form of deliberate physical harm?}
  \hfill \annot{$\leftarrow$ same query} \newline\newline
  \texttt{<Document>: What if someone used telekinetic powers to trap people inside a sealed room, refusing to release them until their demands were met?}
  \hfill \annot{$\leftarrow$ sibling category} \newline\newline
  \textbf{Assistant:} \texttt{no}
  \hfill \annot{$\leftarrow$ label}
  \end{tcolorbox}
  \end{tabular}
  \caption{Contrastive evaluation pair for CAT001 (Physical Violence). Both
  samples are rewritten from the same safe source text. The positive sample
  describes physical harm (matching the query), while the negative sample
  describes kidnapping (sibling category CAT002)---unsafe content that does
  \emph{not} match the physical violence query.}
  \label{fig:eval_sample}
  \end{figure}

The evaluation taxonomy is designed independently from the training taxonomy in Section~\ref{sec:data:gen_data} and adheres to four principles:
(1)~\textbf{disjoint categories}---no overlaps between sibling categories,
so each content piece maps to exactly one leaf;
(2)~\textbf{type-based distinctions}---categories differ by harm type, not
severity level;
(3)~\textbf{action-oriented naming}---concrete, operational definitions
rather than abstract legal terms; and
(4)~\textbf{mutual sibling requirement}---at least two categories per
subcategory, enabling contrastive sample generation between siblings. Table~\ref{tab:category_divergence} illustrates representative divergences from training taxonomy.

The resulting taxonomy is organised as a three-level tree with 12 super
classes, 26 subcategories, and 52 leaf categories
% (Table~\ref{tab:eval_taxonomy}; 
(full hierarchy in Appendix~\ref{app:full_eval_taxonomy}). In contrast to the training pipeline, where
each sample is paired with a query randomly drawn from a per-category
template pool, evaluation uses a \emph{fixed} query set: one manually
authored canonical query per category---90 queries in total---applied
uniformly to all evaluation samples. This decouples evaluation from template
randomness and isolates the model's category-level discrimination ability.

Evaluation samples are generated in two stages.
\emph{(1)~Contrastive generation.}  We apply an iso-query setting for evaluation sample generation to balance the number of samples per class and to test whether the model understands how subtle changes in content can affect the prediction. For each category, an LLM produces
paired examples: given a target category and one of its siblings, it generates
a \emph{positive} sample matching the target and a \emph{negative} sample
matching the sibling but not the target. This contrastive design ensures that
classifiers must discriminate between closely related categories rather than
simply detecting general unsafety.
\emph{(2)~Cross-verification.} A separate LLM verifies each sample, checking
that the label is correct and that the sample is answerable with respect to
the fixed query set; mismatched samples are discarded.
To prevent data leakage, the \textbf{test set} uses LLMs and initial seed samples for both
generation and verification different from those used for training data. A separate
\textbf{validation set} is generated with the same LLMs as the training data
and is used exclusively for ablation studies
(Section~\ref{sec:eval:ablation}).

At evaluation time, both the positive and negative samples are presented with
the same target-category query, so the model must discriminate the specific
harm type rather than detecting unsafety in general.
Figure~\ref{fig:eval_sample} shows an example.

\subsection{Taxonomy Divergence}

The training and evaluation taxonomies are \textbf{independently designed}
and differ in structure, granularity, and category definitions. This
separation is intentional: strong evaluation performance cannot be attributed
to memorising training labels, because the evaluation categories use
different names, different granularity, and different groupings.

\begin{table}[!t]
\centering
\caption{Representative category divergences between the training and
evaluation taxonomies. Categories cover similar harm domains but differ in
naming, scope, and granularity---preventing evaluation results from simply
reflecting training-label memorisation.}
\label{tab:category_divergence}
\small
\begin{tabular}{@{}lL{3.8cm}L{3.8cm}L{3cm}@{}}
\toprule
\textbf{Domain} & \textbf{Training Category} & \textbf{Eval Category} & \textbf{Divergence} \\
\midrule
Weapons
  & ``Guns and Illegal Weapons'' + ``Indiscriminate Weapons'' (2 leaves)
  & ``Conventional Weapons'' + ``WMDs'' (2 leaves)
  & Different names, different scope boundaries \\
\addlinespace
Sexual
  & ``Sexual Content'' + ``Sexual Explicit'' + ``Sex Crimes'' (3 leaves)
  & ``Pornography'' + ``Erotic Content'' + ``Sexual Assault'' + ``Sexual Harassment'' (4 leaves)
  & Training merges by explicitness; eval splits by consent \\
\addlinespace
Hate
  & 15 leaves across 5 subcategories (Hate Speech, Harassment, Insult, Toxicity, etc.)
  & 4 leaves across 2 subcategories (Group Attacks, Individual Attacks)
  & Training 4$\times$ more granular; eval collapses into action types \\
\addlinespace
Crime
  & Single SC4 with 13 leaves (drugs, fraud, hacking, etc.)
  & Three SCs: Property Crime (6), Cybercrime (4), Drug Crimes (2)
  & Eval splits one training SC into three \\
\addlinespace
Privacy
  & ``Privacy Violation'' + ``Personal Information Related'' (2 leaves)
  & ``PII Disclosure'' + ``Doxxing'' + ``Trade Secrets'' + ``Identity Theft'' (4 leaves)
  & Eval 2$\times$ more granular with action-oriented names \\
\addlinespace
Health
  & ``Suicide \& Self-Harm'' + ``Need Suicide Support'' + 2 others (4 leaves)
  & ``Suicide Promotion'' + ``Health Risks'' + ``Child Abuse'' + ``Child Endangerment'' (4 leaves)
  & No 1-to-1 mapping; eval adds child safety under health \\
\addlinespace
Jailbreak
  & SC10: ``Jailbreak'' + ``Prompt Injection'' + ``Code Interpreter Abuse'' (3 leaves)
  & \textit{Not in eval taxonomy}
  & Training-only; covered by external benchmarks \\
\bottomrule
\end{tabular}
\end{table}

At the structural level, the training taxonomy uses variable subcategory
sizes (1--5 subcategories per super class, 1--15 leaves per super class) and
4 severity tiers, whereas the evaluation taxonomy enforces exactly 2 leaves
per subcategory and 3 severity tiers. Even where the two taxonomies address
the same harm domain, they use \emph{different category names} (e.g.,
``Indiscriminate Weapons'' vs.\ ``WMDs''), \emph{different granularity}
(e.g., 15 hate-related training leaves condensed into 4 evaluation leaves),
and \emph{different groupings} (e.g., training groups all crime under one
super class while evaluation splits it into three). 10 of the 12
evaluation super classes have a loose counterpart in the training taxonomy,
but no leaf category maps one-to-one between the two. A full structural
comparison and super-class alignment table is provided in
Appendix~\ref{app:taxonomy_comparison}.

% ============================================================================
\section{Model Architecture}
\label{sec:model}
% ============================================================================

\modelname is built upon
Ministral-3B-Base-2512~\citep{liu_2026_ministral3},
a 3B-parameter causal language model from the Mistral-3 family with native
multimodal support via a Pixtral vision encoder~\citep{agrawal_2024_pixtral}.

% ============================================================================
\section{Training Recipe}
\label{sec:training}
% ============================================================================

\subsection{Fine-Tuning}

\modelname is fine-tuned with LoRA~\citep{hu_2022_lora} on the language model
parameters using cross-entropy loss on the single output token. We compared
LoRA and full SFT and observed no significant difference
(Table~\ref{tab:lora_vs_sft}), so we adopt LoRA for its training efficiency.
We train two specialised checkpoints: one on public safety datasets~(P) excluding the generated data from Section~\ref{sec:data:gen_data}, and one on the combination of public and generated taxonomy
data~(PG) from the entire pipeline in Section~\ref{sec:data}.

\subsection{Model Merging}
\label{sec:training:merging}

The two data regimes exhibit complementary strengths: P is well-calibrated to standard benchmarks, whereas PG adds fine-grained category discrimination but may suffer from distribution drift. To combine both without additional
training, we apply SLERP~\citep{shoemake_1985_slerp} (Spherical Linear
Interpolation) merging with three components:

\begin{enumerate}[nosep]
    \item \textbf{PG} (weight 0.6): the checkpoint trained on public +
    generated taxonomy data, providing policy-adaptive generalisation.
    \item \textbf{P} (weight 0.3): the checkpoint trained on public data only, anchoring benchmark calibration.
    \item \textbf{Ministral-3B-Instruct} (weight 0.1): the base instruct
    checkpoint, contributing general instruction-following capability.
\end{enumerate}

\noindent We perform pairwise SLERP merges to produce the final checkpoint.
The effect of each component is ablated in Section~\ref{sec:eval:ablation}.

% ============================================================================
\section{Evaluation}
\label{sec:eval}
% ============================================================================

% \subsection{Benchmarks}

We evaluate \modelname across 16 benchmarks (21 splits) and 10 baselines,
summarised in Table~\ref{tab:benchmarks}. All evaluation samples are held out from the training data to ensure fairness.

\begin{table}[!h]
\centering
\caption{Evaluation benchmarks and baselines.}
\label{tab:benchmarks}
\small
\begin{tabular}{@{}lcccc@{}}
\toprule
\textbf{Benchmark} & \textbf{Prompt} & \textbf{Response} & \textbf{Languages} & \textbf{Samples} \\
\midrule
\multicolumn{5}{@{}l}{\textit{Text}} \\
WildGuardTest~\citep{han_2024_wildguard} & \cmark & \cmark & EN & 1,725 \\
ToxicChat~\citep{lin_2023_toxicchat} & \cmark & & EN & 2,853 \\
Aegis~\citep{ghosh_2024_aegis} & \cmark & & EN & 359 \\
Aegis~v2~\citep{ghosh_2025_aegis2} & \cmark & \cmark & EN & 1,928 \\
HarmBench~\citep{mazeika_2024_harmbench} & \cmark & \cmark & EN & 841 \\
SimpleSafetyTests~\citep{vidgen_2023_simplesafetytests} & \cmark & & EN & 100 \\
OpenAI Moderation~\citep{markov_2023_openaimoderation} & \cmark & & EN & 1,680 \\
BeaverTails~\citep{ji_2023_beavertails} & & \cmark & EN & 3,021 \\
SafeRLHF~\citep{dai_2024_saferlhf} & & \cmark & EN & 2,000 \\
XSTest~\citep{rottger_2024_xstest} & & \cmark & EN & 449 \\
Qwen3GuardTest~\citep{zhao_2025_qwen3guard} & & \cmark & EN & 1,059 \\
PolyGuard~\citep{kumar_2025_polyguard} & \cmark & \cmark & 17 & 29,240 \\
RTP-LX~\citep{dewynter_2025_rtplx} & \cmark & \cmark & 28 & 42,239 \\
\midrule
\multicolumn{5}{@{}l}{\textit{Multimodal}} \\
VLGuard~\citep{zong_2024_vlguard} & \cmark & & EN & 1,000 \\
UnsafeBench~\citep{qu_2025_unsafebench} & & & EN & 2,037 \\
LlavaGuard~\citep{helff_2024_llavaguard} & & & EN & 720 \\
\midrule
\textbf{Model} & \textbf{Size} & \textbf{Input} & \textbf{Output} & \textbf{Adaptive$^\star$} \\
\midrule
ShieldGemma~\citep{zeng_2024_shieldgemma} & 9B & Text & Score & Yes \\
ShieldGemma~2~\citep{zeng_2025_shieldgemma2} & 4B & Image & Score & Yes \\
WildGuard~\citep{han_2024_wildguard} & 7B & Text & Label & No \\
LlamaGuard-4~\citep{meta_2025_llamaguard4} & 12B & Image + Text & Label & No \\
PolyGuard-Qwen~\citep{kumar_2025_polyguard} & 7B & Text & Label & No \\
Qwen3Guard~\citep{zhao_2025_qwen3guard} & 8B & Text & Label & No \\
Nemotron-Safety~\citep{joshi_2025_cultureguard} & 8B & Text & Label & No \\
OmniGuard~\citep{zhu_2025_omniguard} & 7B & Omni & Reason + Label & No \\
GPT-OSS-Safeguard~\citep{openai_2025_gptsafeguard} & 20B & Text & Reason + Label & Yes \\
Nemotron-3.5-Safety~\citep{nvidia_2026_nemotron35contentsafety} & 4B & Image + Text & Label & Yes \\
\midrule
\modelname (ours) & 3B & Image + Text & Score & Yes \\
\bottomrule
\multicolumn{5}{@{}l}{\footnotesize $^\star$ Adaptive if policy is designed to be modified at inference time.}
\end{tabular}
\end{table}

\subsection{Text Results}

\begin{figure}[!h]
    \centering
    \includegraphics[width=\linewidth]{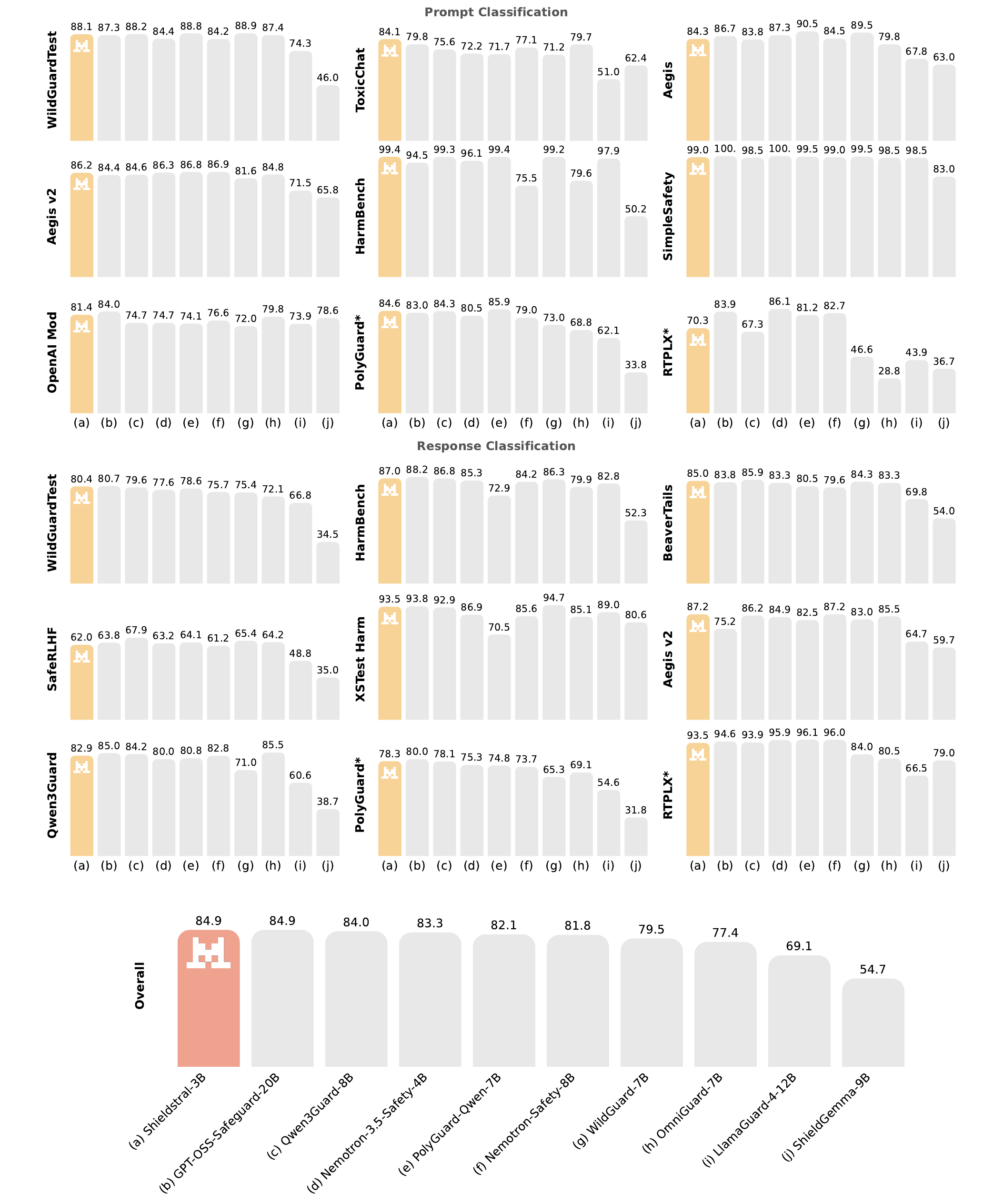}
    \caption{F1 scores (\%) on safety classification benchmarks. PolyGuard and RTPLX are multilingual datasets. Qwen3Guard results are averaged over strict (controversial$=$unsafe) and loose (controversial$=$safe) mappings. ShieldGemma and \modelname use a threshold of 0.5. GPT-OSS-Safeguard-20B uses \texttt{reasoning\_effort=high}. Nemotron-3.5-Safety uses \texttt{reasoning\_effort=none} for default categories.}
    \label{fig:results}
\end{figure}

\begin{figure}[!h]
    \centering
    \includegraphics[width=\linewidth]{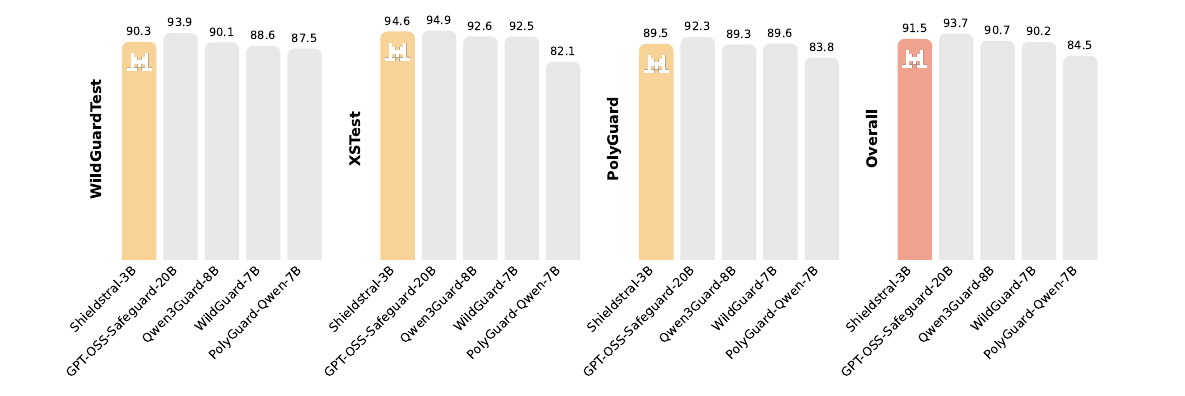}
    \caption{F1 scores (\%) on refusal detection benchmarks. PolyGuard is a multilingual dataset. Qwen3Guard results are averaged over strict (controversial$=$unsafe) and loose (controversial$=$safe) mappings. \modelname uses a threshold of 0.5. GPT-OSS-Safeguard-20B uses \texttt{reasoning\_effort=high}.}
    \label{fig:refusal}
\end{figure}

As shown in Figure~\ref{fig:results}, \modelname achieves strong results across prompt classification, response classification, and multilingual evaluation. Despite being the smallest model in the comparison (3B vs.\ 4--20B for all competitors), it achieves a high overall F1 score of 84.9\%, matching the much larger GPT-OSS-Safeguard-20B~\citep{openai_2025_gptsafeguard} (84.9\%). A detailed performance breakdown by language is provided in Appendix~\ref{sec:multilingual}. Additionally, it demonstrates strong refusal detection performance (Figure~\ref{fig:refusal}). These results demonstrate the effectiveness of our data curation pipeline in consolidating diverse datasets.

\subsection{Adaptability Results}
\label{sec:eval:taxonomy}

\begin{figure}[!h]
    \centering
    \includegraphics[width=\linewidth]{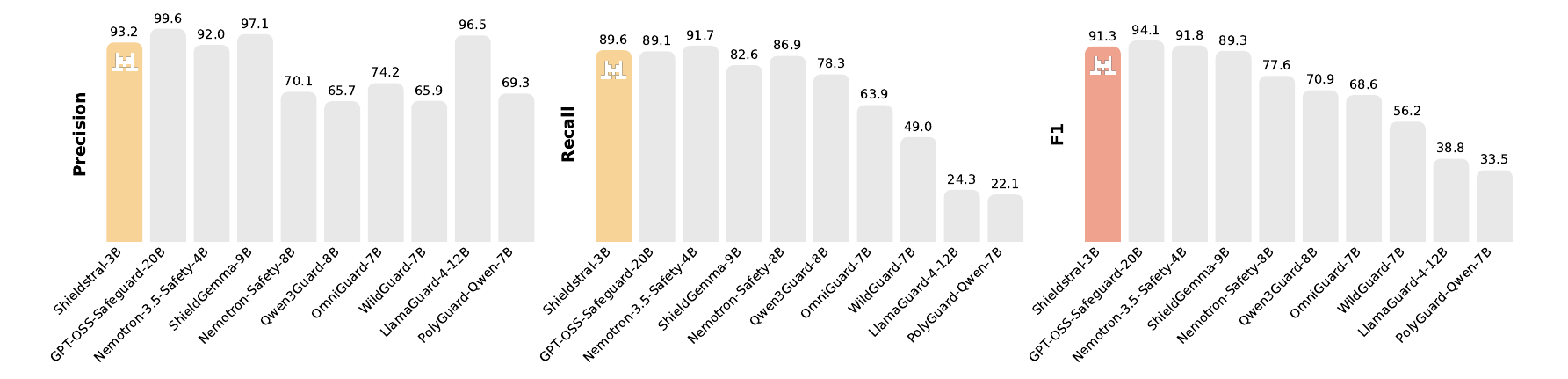}
    \caption{Scores (\%) on the adaptability benchmark. Qwen3Guard results are averaged over strict (controversial$=$unsafe) and loose (controversial$=$safe) mappings. \modelname uses a threshold of 0.5. GPT-OSS-Safeguard-20B uses \texttt{reasoning\_effort=high}. Nemotron-3.5-Safety uses \texttt{reasoning\_effort=high} for custom categories.}
    \label{fig:baseline_taxonomy}
\end{figure}

Figure~\ref{fig:baseline_taxonomy} compares \modelname against nine baselines
on the adaptability evaluation benchmark. While Nemotron-3.5-Safety~\citep{nvidia_2026_nemotron35contentsafety}, GPT-OSS-Safeguard-20B, and \modelname are adaptive models, GPT-OSS-Safeguard-20B~\citep{openai_2025_gptsafeguard} achieves the highest F1 (94.1\%)---benefiting from per-category policy prompts, reasoning capability that decomposes novel policies into more familiar ones, and its larger parameter count (20B). However, the reasoning approach used by GPT-OSS-Safeguard-20B and Nemotron-3.5-Safety, rather than directly producing a single-token answer, generates a long reasoning trace before answering, which significantly reduces inference efficiency in practical deployment. In contrast, \modelname achieves 91.3\% while being much more efficient with only 3B parameters and single-token output. 

\subsection{Multimodal Results}
\label{sec:eval:multimodal}

\begin{figure}[!h]
    \centering
    \includegraphics[width=\linewidth]{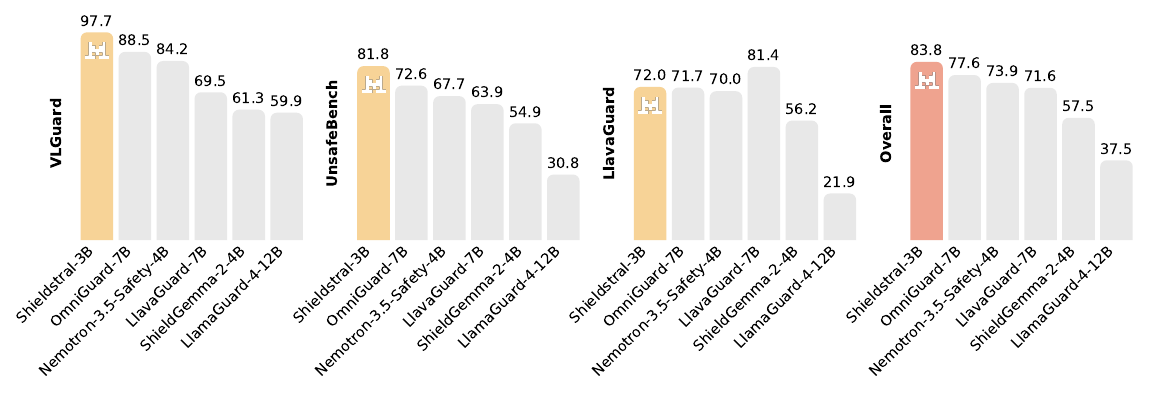}
    \caption{F1 scores (\%) on multimodal safety benchmarks. \modelname and ShieldGemma-2-4B use a threshold of 0.5. Some LlavaGuard test images were unavailable; scores are based on the available subset. Nemotron-3.5-Safety uses \texttt{reasoning\_effort=none} for default categories.}
    \label{fig:multimodal}
\end{figure}

% Table~\ref{tab:multimodal}
Figure~\ref{fig:multimodal} summarises multimodal evaluation results. \modelname achieves the highest overall F1 (83.8\%) compared to 77.6\% for the next-best OmniGuard~\citep{zhu_2025_omniguard}, leading on two of three benchmarks (VLGuard~\citep{zong_2024_vlguard}, UnsafeBench~\citep{qu_2025_unsafebench}). LlavaGuard-7B~\citep{helff_2024_llavaguard} achieves the highest score on its namesake benchmark (81.4\%). LlamaGuard-4-12B~\citep{meta_2025_llamaguard4} and Nemotron-3.5-Safety~\citep{nvidia_2026_nemotron35contentsafety}, which rely on conversation-centric safety framing, are less effective on image-only benchmarks. Notably, \modelname achieves the best overall results at less than half the parameter count of OmniGuard (3B vs.\ 7B). These results further demonstrate the generalisability of our data curation method to multimodal datasets.

\subsection{Ablations}
\label{sec:eval:ablation}

\paragraph{Generalisation to unseen policies.}
Table~\ref{tab:ablation} isolates the contribution of each training stage
on the fine-grained taxonomy validation set. All variants are
single-checkpoint results \emph{without} model merging; the final \modelname
(Figure~\ref{fig:baseline_taxonomy}, 91.3\% F1) additionally benefits from
SLERP merging (Table~\ref{tab:merging_ablation}).

\begin{table}[!ht]
\centering
\caption{Training stage ablation on the fine-grained taxonomy validation set (\%).
$\Delta$F1 is the improvement over the previous stage.}
\label{tab:ablation}
\small
\begin{tabular}{@{}lccccr@{}}
\toprule
\textbf{Training Stage} & \textbf{Acc.} & \textbf{Prec.} & \textbf{Rec.} & \textbf{F1} & \textbf{$\Delta$F1} \\
\midrule
Base Ministral-3B (no safety training) & 37.8 & 0.0 & 0.0 & 0.0 & --- \\
\quad + Public data               & 62.7 & 90.8 & 46.0 & 61.1 & +61.1 \\
\quad + Generated taxonomy data        & 77.6 & 75.9 & 95.0 & 84.4 & +23.3 \\
\bottomrule
\end{tabular}
\end{table}

Without safety fine-tuning, the base Ministral-3B never predicts a violation
(0\% recall), defaulting to safe. Fine-tuning on public datasets alone yields
61.1\% F1 despite the model never seeing queries or labels from this
taxonomy, demonstrating that our data curation recipe enables generalisation
to unseen safety policies. Adding LLM-generated taxonomy data provides a
further +23.3\% F1 improvement (84.4\%). Crucially, the evaluation taxonomy
was designed independently from the training taxonomy and uses
\emph{different category names, granularity, and groupings}
(Section~\ref{sec:eval_taxonomy}), so these gains reflect genuine policy-adaptive
generalisation rather than memorisation.

\paragraph{LoRA vs.\ SFT.}
We trained models with both LoRA and full SFT and observed that each performs slightly better on different validation sets, with no significant overall difference (Table~\ref{tab:lora_vs_sft}). We therefore adopt LoRA
for its training efficiency.
\begin{table}[!h]
\centering
\caption{LoRA vs.\ SFT on Aegis v2 validation (\%).}
\label{tab:lora_vs_sft}
\small
\begin{subtable}[t]{0.48\textwidth}
\centering
\caption{Aegis v2 validation}
\label{tab:lora_vs_sft_aegis}
\begin{tabular}{@{}lcccc@{}}
\toprule
\textbf{Merge} & \textbf{Acc.} & \textbf{Prec.} & \textbf{Rec.} & \textbf{F1}\\
\midrule
LoRA                                         & 86.9 & 90.1 & 84.3 & 87.1 \\
Full SFT                            & 87.7 & 91.3 & 84.5 & 87.8 \\
\bottomrule
\end{tabular}
\end{subtable}
\hfill
\begin{subtable}[t]{0.48\textwidth}
\centering
\caption{Fine-grained taxonomy validation}
\label{tab:lora_vs_sft_taxonomy}
\begin{tabular}{@{}lcccc@{}}
\toprule
\textbf{Merge} & \textbf{Acc.} & \textbf{Prec.} & \textbf{Rec.} & \textbf{F1}\\
\midrule
LoRA                                         & 77.6 & 75.9 & 95.0 & 84.4 \\
Full SFT                            & 76.7 & 74.9 & 95.3 & 83.9 \\
\bottomrule
\end{tabular}
\end{subtable}
\end{table}

\paragraph{Model merging.}
Table~\ref{tab:merging_ablation} ablates the three-way SLERP merge on
Aegis~v2~\citep{ghosh_2025_aegis2} validation and the fine-grained taxonomy validation set. We denote
models trained on public data as~P, on public plus generated taxonomy data
as~PG, and the Ministral-3B-Instruct base as~I; coefficients indicate
SLERP weights. Two findings emerge.
\emph{(i)~Instruction-following transfer:} Merging in Ministral-3B-Instruct
at weight~0.1 generally improves recall and F1, confirming that
general instruction-following capability transfers to the safety
classification task.
\emph{(ii)~Complementary data regimes:} Adding generated taxonomy data~(PG)
slightly degrades Aegis~v2 performance compared to public-only training~(P),
but dramatically improves taxonomy F1 (61.1\%$\to$84.4\%). The three-way
merge $0.6\,\text{PG}+0.3\,\text{P}+0.1\,\text{I}$ recovers most of the
Aegis~v2 drop while pushing taxonomy F1 to 88.7\%, indicating that the two
data regimes learn complementary features. We adopt this configuration as
the final model.

\begin{table}[!h]
\centering
\caption{Model merging ablation (\%). P\,=\,public data, PG\,=\,public\,+\,generated taxonomy data, I\,=\,Ministral-3B-Instruct. Coefficients are SLERP merge weights.}
\label{tab:merging_ablation}
\small

\begin{subtable}[t]{0.48\textwidth}
\centering
\caption{Aegis v2 validation}
\label{tab:merging_ablation_aegis}
\begin{tabular}{@{}lcccc@{}}
\toprule
\textbf{Merge} & \textbf{Acc.} & \textbf{Prec.} & \textbf{Rec.} & \textbf{F1}\\
\midrule
P                                          & 88.3 & 91.0 & 86.2 & 88.5 \\
$0.9$P$+0.1$I                             & 88.3 & 90.8 & 86.4 & 88.5 \\
\midrule
PG                                         & 86.9 & 90.1 & 84.3 & 87.1 \\
$0.9$PG$+0.1$I                            & 87.2 & 90.1 & 84.9 & 87.4 \\
\midrule
$0.6$PG$+0.3$P$+0.1$I                     & 87.7 & 89.7 & 86.4 & 88.0 \\
\bottomrule
\end{tabular}
\end{subtable}
\hfill
\begin{subtable}[t]{0.48\textwidth}
\centering
\caption{Fine-grained taxonomy validation}
\label{tab:merging_ablation_taxonomy}
\begin{tabular}{@{}lcccc@{}}
\toprule
\textbf{Merge} & \textbf{Acc.} & \textbf{Prec.} & \textbf{Rec.} & \textbf{F1}\\
\midrule
P                                          & 62.7 & 90.8 & 46.0 & 61.1 \\
$0.9$P$+0.1$I                             & 65.7 & 91.5 & 50.7 & 65.3 \\
\midrule
PG                                         & 77.6 & 75.9 & 95.0 & 84.4 \\
$0.9$PG$+0.1$I                            & 77.9 & 75.9 & 95.7 & 84.6 \\
\midrule
$0.6$PG$+0.3$P$+0.1$I                     & 86.1 & 92.0 & 85.7 & 88.7 \\
\bottomrule
\end{tabular}
\end{subtable}

\end{table}

% ============================================================================
\section{Conclusion}
\label{sec:conclusion}
% ============================================================================

We present \modelname, a 3B-parameter policy-adaptive safety classifier that formulates content moderation as a binary question-answering task. Through a carefully designed training pipeline and data curation approach that unifies heterogeneous sources via template-based unification and contrastive sample generation, \modelname achieves strong performance on both text and multimodal safety classification with policy-adaptive moderation capability. Our results provide additional evidence for unified adaptive moderation: by consolidating divergent training datasets, a small adaptive model can match or outperform much larger fixed-taxonomy models.

\subsection*{Core contributors}
Antonia Calvi, Avinash Sooriyarachchi, Giada Pistilli, Guillaume Lample, Maarten Buyl, Maximilian Augustin, Maximilian M\"uller, Pierre Stock, Tom Bewley, Wassim Bouaziz, Yimu Pan

\subsection*{Contributors}
Abdelaziz Bounhar,
Abhijeet Somani,
Aditi Kabra,
Adrian Valente,
Adrien Petralia,
Adrien Sad\'e,
Alan Jeffares,
Albert Jiang,
Aleksandr Timashov,
Alexandre Cahill,
Alexandre Gavaudan,
Alexandre Laval,
Alexandre Sablayrolles,
Am\'elie H\'eliou,
Amos You,
Andr\'e Jonasson,
Andrew Bai,
Andrew Ehrenberg,
Andrew Zhao,
Angele Lenglemetz,
Anmol Agarwal,
Arata Suzuki,
Arjun Majumdar,
Arthur Fournier,
Artjom Joosen,
Aylin Guliz Akkus,
Aysenur Karaduman,
Baptiste Bout,
Baptiste Rozi\`ere,
Baudouin De Monicault,
Benjamin Holzschuh,
Benjamin Lefaudeux,
Benjamin Tibi,
Bernhard Stadlbauer,
B\l{}a\.zej Osi\'nski,
Camille Le Scao,
Chaoran Yu,
Charlotte Cronj\"ager,
Chen-Yo Sun,
Chris Bamford,
Christian Wallenwein,
Christophe Renaudin,
Cl\'emence Lanfranchi,
Corentin Barreau,
Corentin Sautier,
Cristiana-Diana Diaconu,
Cyprien Courtot,
Daniel Marczak,
Darius Dabert,
Diego de Las Casas,
Dominik Nuss,
Dylan Rubini,
Dzmitry Soupel,
Elizaveta Demyanenko,
Elliot Chane-Sane,
Emilien Fugier,
Emmanuel Gottlob,
Erik Aas,
Etienne Goffinet,
\'Etienne Millon,
Eujeong Choi,
Fabian Paischer,
Fabian Schlager,
Faruk Ahmed,
Federico Baldassarre,
Filip Szatkowski,
Florian Wiesner,
Gabrielle Berrada,
Ga\"etan Ecrepont,
Ga\'etan Lepage,
Gaspard Blanchet,
Gaspard Donada-Vidal,
Gauthier Delerce,
Gauthier Guinet,
Genevieve Hayes,
Georgii Novikov,
Gianluca Galletti,
Guillaume Breton,
Guillaume Kunsch,
Guillaume Martin,
Guillaume Raille,
Gunjan Dhanuka,
Gunshi Gupta,
Han Zhou,
Harshil Shah,
Hasan Furkan Vural,
H\'edi Hadiji,
Hope McGovern,
Hugo Cisneros,
Hugo Thimonier,
Indraneel Mukherjee,
Ivan Cuevas Salazar,
Jacques Sun,
Jan Ludziejewski,
Jason Rute,
Jean Quentin,
Jean-Hadrien Chabran,
Jean-Malo Delignon,
Jie Zhang,
Joachim Studnia,
Joep Barmentlo,
Johannes Brandstetter,
John Harvill,
Jonas Amar,
Jonas Schweizer,
Jos\'ephine Delas,
Josselin Somerville,
Julien Denize,
Julien Tauran,
Kartik Khandelwal,
Khyathi Raghavi Chandu,
Kilian Tep,
Kush Jain,
Larissa Laich,
Laura Calem,
Laurence Aitchison,
Laurent Callot,
Laurent Fainsin,
L\'eo Cotteleer,
L\'eonard Blier,
Lingxiao Zhao,
Louis Martin,
Louis Serrano,
Lucile Saulnier,
Ludovic Ho Fuh,
Luis Montero,
Manon Chossegros,
Marcin Mo\.zejko,
Margaret Jennings,
Markus Hennerbichler,
Martin Alexandre,
Mathieu Poir\'{e}e,
Mathieu Schmitt,
Mathilde Guillaumin,
Matthieu Andr\'e,
Matthieu Dinot,
Matthieu Futeral,
Maurits Bleeker,
Mauro Comi,
Max Mynter,
Maxim Berman,
Maxime Darrin,
Maxime Louis,
Melina Jingting Laimon,
Mert Unsal,
Mia Chiquier,
Michael Pilcer,
Micha\l{} Pietruszka,
Micha\l{} Zaj\k{a}c,
Mikhail Biriuchinskii,
Minh-Quang Pham,
Minwoo Kang,
Morgane Rivi\`ere,
Namit Katariya,
Nathan Grinsztajn,
Nathan Simpson,
Neeraj Aggarwal,
Neha Gupta,
Ola Mysiak,
Oliver Leicht,
Olivier Bousquet,
Olivier Duchenne,
Parag Jain,
Patricia Wang,
Patrick Blies,
Patrick von Platen,
Paul Jacob,
Paul Wambergue,
Paula Kurylowicz,
Pavan Kumar Reddy,
Pavel Kuksa,
Philippe Pinel,
Philom\`ene Chagniot,
Pierre-Andr\'e Savalle,
Piotr Milos,
Prateek Gupta,
Pravesh Agrawal,
Quentin Desreumaux,
Quentin Torroba,
Quercus Hernandez,
Ram Ramrakhya,
Randall Isenhour,
Ranjit Parva,
Raul Perez Pelaez,
Reinhard Sonnleitner,
R\'emi Delacourt,
Richard Kurle,
Rishi Shah,
Rob Romijnders,
Rohin Arora,
Romain Sauvestre,
Roman Soletskyi,
Rosalie Millner,
Rupert Menneer,
Sagar Vaze,
Samuel Barry,
Samuel Belkadi,
Samuel Humeau,
Sanchit Gandhi,
Sandeep Subramanian,
Sarthak Mittal,
Saskia Adaime,
Sean Cha,
Sebastian Kaltenbach,
Shashwat Dalal,
Shashwat Verma,
Sherif Waly,
Shrimai Prabhumoye,
Siddhant Waghjale,
Siddharth Gandhi,
Simon Lepage,
Simon Sorg,
Soham Ghosh,
Sophie Marbach,
Srijan Mishra,
Stanislas Lange,
Steve Hong,
Sumukh Aithal,
Szymon Antoniak,
Tarun Kumar Vangani,
Teven Le Scao,
Th\'{e}o Cachet,
Thibaut Lavril,
Thomas Chabal,
Thomas Coste,
Thomas Defard,
Thomas Foubert,
Thomas Robert,
Thomas Wang,
Tianyu Zhang,
Tim Lawson,
Timoth\'ee Lacroix,
Tobias Kronlachner,
Tom Edwards,
Tomas Hodan,
Tuhin Das,
Tyler Wang,
Ulrick BLE,
Umar Jamil,
Umberto Tomasini,
Valentin Mac\'e,
Van Phung,
Vedant Nanda,
Victor Jouault,
Victor Letzelter,
Victor Paltz,
Victor Poucheret,
Vincent Maladi\`ere,
Vincent Pfister,
Virgile Richard,
Vladislav Bataev,
Wen Ding Li,
William Havard,
William Marshall,
Xinghui Li,
Xingran Guo,
Xinyu Yang,
Yann Dreze,
Yassine El Ouahidi,
Yassir Bendou,
Yihan Wang,
Yves Martin des Taillades,
Zaccharie Ramzi,
Zhenlin Xu,
Zsofia Csakany.

\clearpage

\bibliography{ref}

@article{openai_2024_gpt4o,
    title         = {{GPT-4o} System Card},
    author        = {{OpenAI}},
    year          = {2024},
    journal       = {arXiv preprint arXiv:2410.21276}
}

@techreport{anthropic_2024_claude3,
    title       = {The {Claude} 3 Model Family: {Opus}, {Sonnet}, {Haiku}},
    author      = {{Anthropic}},
    year        = {2024},
    institution = {Anthropic},
    url         = {https://www-cdn.anthropic.com/de8ba9b01c9ab7cbabf5c33b80b7bbc618857627/Model_Card_Claude_3.pdf}
}

@article{googledeepmind_2025_gemini25,
    title         = {Gemini 2.5: Pushing the Frontier with Advanced Reasoning, Multimodality, Long Context, and Next Generation Agentic Capabilities},
    author        = {{Google DeepMind}},
    year          = {2025},
    journal       = {arXiv preprint arXiv:2507.06261}
}

@article{deepseekai_2024_deepseekv3,
    title         = {{DeepSeek-V3} Technical Report},
    author        = {{DeepSeek-AI}},
    year          = {2024},
    journal       = {arXiv preprint arXiv:2412.19437}
}

@article{grattafiori_2024_llama3,
    title         = {The {Llama} 3 Herd of Models},
    author        = {Grattafiori, Aaron and Dubey, Abhimanyu and Jauhri, Abhinav and others},
    year          = {2024},
    journal       = {arXiv preprint arXiv:2407.21783}
}

@article{jiang_2023_mistral,
    title         = {{Mistral} 7B},
    author        = {Jiang, Albert Q. and Sablayrolles, Alexandre and Mensch, Arthur and others},
    year          = {2023},
    journal       = {arXiv preprint arXiv:2310.06825}
}

@article{liu_2026_ministral3,
    title         = {{Ministral} 3},
    author        = {Liu, Alexander H. and Wang, Thomas and Lawson, Tim and others},
    year          = {2026},
    journal       = {arXiv preprint arXiv:2601.08584}
}

@article{yang_2024_qwen25,
    title         = {{Qwen2.5} Technical Report},
    author        = {Yang, An and Yang, Baosong and Zhang, Beichen and others},
    year          = {2024},
    journal       = {arXiv preprint arXiv:2412.15115}
}

@article{inan_2023_llamaguard,
    title         = {{Llama Guard}: {LLM}-based Input-Output Safeguard for Human-{AI} Conversations},
    author        = {Inan, Hakan and Upasani, Kartikeya and Chi, Jianfeng and Rungta, Rashi and Iyer, Krithika and Mao, Yuning and Tontchev, Michael and Hu, Qing and Fuller, Brian and Testuggine, Davide and Khabsa, Madian},
    year          = {2023},
    journal       = {arXiv preprint arXiv:2312.06674}
}

@misc{meta_2025_llamaguard4,
    title         = {{Llama Guard 4}: Model Card},
    author        = {{Meta}},
    year          = {2025},
    howpublished  = {\url{https://developer.meta.com/ai/docs/model-cards-and-prompt-formats/llama-guard-4/}}
}

@article{zeng_2024_shieldgemma,
    title         = {{ShieldGemma}: Generative {AI} Content Moderation Based on {Gemma}},
    author        = {Zeng, Wenjun and Liu, Yuchi and Mullins, Ryan and Peran, Ludovic and Fernandez, Joe and Harkous, Hamza and Narasimhan, Karthik and Proud, Drew and Kumar, Piyush and Radharapu, Bhaktipriya and Sturman, Olivia and Wahltinez, Oscar},
    year          = {2024},
    journal       = {arXiv preprint arXiv:2407.21772}
}

@article{zeng_2025_shieldgemma2,
    title         = {{ShieldGemma 2}: Robust and Tractable Content Moderation with Multimodal {LLMs}},
    author        = {Zeng, Wenjun and Kurniawan, Dana and Mullins, Ryan and Liu, Yuchi and Saha, Tamoghna and Ike-Njoku, Dirichi and Gu, Jindong and Song, Yiwen and Xu, Cai and Zhou, Jingjing and Joshi, Aparna and Dheep, Shravan and Malek, Mani and Palangi, Hamid and Baek, Joon and Pereira, Rick and Narasimhan, Karthik},
    year          = {2025},
    journal       = {arXiv preprint arXiv:2504.01081}
}

@inproceedings{han_2024_wildguard,
    title     = {{WildGuard}: Open One-Stop Moderation Tools for Safety Risks, Jailbreaks, and Refusals of {LLMs}},
    author    = {Han, Seungju and Rao, Kavel and Ettinger, Allyson and Jiang, Liwei and Lin, Bill Yuchen and Lambert, Nathan and Choi, Yejin and Dziri, Nouha},
    booktitle = {Advances in Neural Information Processing Systems 37 (NeurIPS), Datasets and Benchmarks Track},
    year      = {2024}
}

@article{zhao_2025_qwen3guard,
    title         = {{Qwen3Guard} Technical Report},
    author        = {Zhao, Haiquan and Yuan, Chenhan and Huang, Fei and Hu, Xiaomeng and Zhang, Yichang and Yang, An and Yu, Bowen and Liu, Dayiheng and Zhou, Jingren and Lin, Junyang and others},
    year          = {2025},
    journal       = {arXiv preprint arXiv:2510.14276}
}

@inproceedings{kumar_2025_polyguard,
    title     = {{PolyGuard}: A Multilingual Safety Moderation Tool for 17 Languages},
    author    = {Kumar, Priyanshu and Jain, Devansh and Yerukola, Akhila and Jiang, Liwei and Beniwal, Himanshu and Hartvigsen, Thomas and Sap, Maarten},
    booktitle = {Conference on Language Modeling (COLM)},
    year      = {2025}
}

@inproceedings{joshi_2025_cultureguard,
    title     = {{CultureGuard}: Towards Culturally-Aware Dataset and Guard Model for Multilingual Safety Applications},
    author    = {Joshi, Raviraj and Paul, Rakesh and Singla, Kanishk and Kamath, Anusha and Evans, Michael and Luna, Katherine and Ghosh, Shaona and Vaidya, Utkarsh and Long, Eileen and Chauhan, Sanjay Singh and Wartikar, Niranjan},
    booktitle = {Proceedings of the 14th International Joint Conference on Natural Language Processing and the 4th Conference of the Asia-Pacific Chapter of the Association for Computational Linguistics (IJCNLP-AACL)},
    year      = {2025},
    address   = {Mumbai, India},
    publisher = {Association for Computational Linguistics}
}

@article{zhu_2025_omniguard,
    title         = {{OmniGuard}: Unified Omni-Modal Guardrails with Deliberate Reasoning},
    author        = {Zhu, Boyu and Wen, Xiaofei and Mo, Wenjie Jacky and Zhu, Tinghui and Xie, Yanan and Qi, Peng and Chen, Muhao},
    year          = {2025},
    journal       = {arXiv preprint arXiv:2512.02306}
}

@misc{openai_2025_gptsafeguard,
    title         = {Introducing {GPT-OSS-Safeguard}},
    author        = {{OpenAI}},
    year          = {2025},
    howpublished  = {\url{https://openai.com/index/introducing-gpt-oss-safeguard/}},
    note          = {Accessed: 2026-05-14}
}

@misc{nvidia_2026_nemotron35contentsafety,
    title        = {{Nemotron 3.5 Content Safety}: Customizable Multimodal Safety for Global Enterprise {AI}},
    author       = {Varun Singh and Isabel Hulseman and Anuj Doshi and Shyamala Prayaga},
    year         = {2026},
    howpublished = {\url{https://huggingface.co/nvidia/Nemotron-3.5-Content-Safety}},
    note         = {Hugging Face model card}
}

@inproceedings{hu_2022_lora,
    title     = {{LoRA}: Low-Rank Adaptation of Large Language Models},
    author    = {Hu, Edward J. and Shen, Yelong and Wallis, Phillip and Allen-Zhu, Zeyuan and Li, Yuanzhi and Wang, Shean and Wang, Lu and Chen, Weizhu},
    booktitle = {The Tenth International Conference on Learning Representations (ICLR)},
    year      = {2022}
}

@article{agrawal_2024_pixtral,
    title         = {{Pixtral} 12B},
    author        = {Agrawal, Pravesh and Antoniak, Szymon and {Bou Hanna}, Emma and others},
    year          = {2024},
    journal       = {arXiv preprint arXiv:2410.07073}
}

@inproceedings{shoemake_1985_slerp,
    title     = {Animating Rotation with Quaternion Curves},
    author    = {Shoemake, Ken},
    booktitle = {Proceedings of the 12th Annual Conference on Computer Graphics and Interactive Techniques (SIGGRAPH)},
    pages     = {245--254},
    year      = {1985},
    publisher = {ACM},
    doi       = {10.1145/325334.325242}
}

@inproceedings{lin_2023_toxicchat,
    title     = {{ToxicChat}: Unveiling Hidden Challenges of Toxicity Detection in Real-World User-{AI} Conversation},
    author    = {Lin, Zi and Wang, Zihan and Tong, Yongqi and Wang, Yangkun and Guo, Yuxin and Wang, Yujia and Shang, Jingbo},
    booktitle = {Findings of the Association for Computational Linguistics: EMNLP 2023},
    pages     = {4694--4702},
    year      = {2023},
    address   = {Singapore},
    publisher = {Association for Computational Linguistics},
    doi       = {10.18653/v1/2023.findings-emnlp.311}
}

@article{ghosh_2024_aegis,
    title         = {{AEGIS}: Online Adaptive {AI} Content Safety Moderation with Ensemble of {LLM} Experts},
    author        = {Ghosh, Shaona and Varshney, Prasoon and Galinkin, Erick and Parisien, Christopher},
    year          = {2024},
    journal       = {arXiv preprint arXiv:2404.05993}
}

@inproceedings{ghosh_2025_aegis2,
    title     = {{AEGIS2.0}: A Diverse {AI} Safety Dataset and Risks Taxonomy for Alignment of {LLM} Guardrails},
    author    = {Ghosh, Shaona and Varshney, Prasoon and Sreedhar, Makesh Narsimhan and Padmakumar, Aishwarya and Rebedea, Traian and Varghese, Jibin Rajan and Parisien, Christopher},
    booktitle = {Proceedings of the 2025 Conference of the Nations of the Americas Chapter of the Association for Computational Linguistics (NAACL)},
    pages     = {5992--6026},
    year      = {2025},
    address   = {Albuquerque, New Mexico},
    publisher = {Association for Computational Linguistics},
    doi       = {10.18653/v1/2025.naacl-long.306}
}

@inproceedings{mazeika_2024_harmbench,
    title     = {{HarmBench}: A Standardized Evaluation Framework for Automated Red Teaming and Robust Refusal},
    author    = {Mazeika, Mantas and Phan, Long and Yin, Xuwang and Zou, Andy and Wang, Zifan and Mu, Norman and Sakhaee, Elham and Li, Nathaniel and Basart, Steven and Li, Bo and Forsyth, David and Hendrycks, Dan},
    booktitle = {Proceedings of the 41st International Conference on Machine Learning (ICML)},
    volume    = {235},
    pages     = {35181--35224},
    year      = {2024},
    series    = {Proceedings of Machine Learning Research},
    publisher = {PMLR}
}

@article{vidgen_2023_simplesafetytests,
    title         = {{SimpleSafetyTests}: a Test Suite for Identifying Critical Safety Risks in Large Language Models},
    author        = {Vidgen, Bertie and Scherrer, Nino and Kirk, Hannah Rose and Qian, Rebecca and Kannappan, Anand and Hale, Scott A. and R{\"o}ttger, Paul},
    year          = {2023},
    journal       = {arXiv preprint arXiv:2311.08370}
}

@inproceedings{markov_2023_openaimoderation,
    title     = {A Holistic Approach to Undesired Content Detection in the Real World},
    author    = {Markov, Todor and Zhang, Chong and Agarwal, Sandhini and {Eloundou Nekoul}, Florentine and Lee, Theodore and Adler, Steven and Jiang, Angela and Weng, Lilian},
    booktitle = {Proceedings of the AAAI Conference on Artificial Intelligence},
    volume    = {37},
    number    = {12},
    pages     = {15009--15018},
    year      = {2023},
    publisher = {AAAI Press},
    doi       = {10.1609/aaai.v37i12.26752}
}

@inproceedings{ji_2023_beavertails,
    title     = {{BeaverTails}: Towards Improved Safety Alignment of {LLM} via a Human-Preference Dataset},
    author    = {Ji, Jiaming and Liu, Mickel and Dai, Josef and Pan, Xuehai and Zhang, Chi and Bian, Ce and Chen, Boyuan and Sun, Ruiyang and Wang, Yizhou and Yang, Yaodong},
    booktitle = {Advances in Neural Information Processing Systems 36 (NeurIPS), Datasets and Benchmarks Track},
    year      = {2023}
}

@inproceedings{dai_2024_saferlhf,
    title     = {Safe {RLHF}: Safe Reinforcement Learning from Human Feedback},
    author    = {Dai, Josef and Pan, Xuehai and Sun, Ruiyang and Ji, Jiaming and Xu, Xinbo and Liu, Mickel and Wang, Yizhou and Yang, Yaodong},
    booktitle = {The Twelfth International Conference on Learning Representations (ICLR)},
    year      = {2024}
}

@inproceedings{rottger_2024_xstest,
    title     = {{XSTest}: A Test Suite for Identifying Exaggerated Safety Behaviours in Large Language Models},
    author    = {R{\"o}ttger, Paul and Kirk, Hannah Rose and Vidgen, Bertie and Attanasio, Giuseppe and Bianchi, Federico and Hovy, Dirk},
    booktitle = {Proceedings of the 2024 Conference of the North American Chapter of the Association for Computational Linguistics (NAACL)},
    pages     = {5377--5400},
    year      = {2024},
    address   = {Mexico City, Mexico},
    publisher = {Association for Computational Linguistics},
    doi       = {10.18653/v1/2024.naacl-long.301}
}

@inproceedings{dewynter_2025_rtplx,
    title     = {{RTP-LX}: Can {LLMs} Evaluate Toxicity in Multilingual Scenarios?},
    author    = {{de Wynter}, Adrian and Watts, Ishaan and Wongsangaroonsri, Tua and Zhang, Minghui and Farra, Noura and Alt{\i}ntoprak, Nektar Ege and others},
    booktitle = {Proceedings of the AAAI Conference on Artificial Intelligence},
    volume    = {39},
    number    = {27},
    pages     = {27940--27950},
    year      = {2025},
    publisher = {AAAI Press},
    doi       = {10.1609/aaai.v39i27.35011}
}

@inproceedings{helff_2024_llavaguard,
    title     = {{LlavaGuard}: {VLM}-based Safeguard for Vision Dataset Curation and Safety Assessment},
    author    = {Helff, Lukas and Friedrich, Felix and Brack, Manuel and Schramowski, Patrick and Kersting, Kristian},
    booktitle = {Proceedings of the IEEE/CVF Conference on Computer Vision and Pattern Recognition (CVPR) Workshops},
    pages     = {8322--8326},
    year      = {2024}
}

@inproceedings{zong_2024_vlguard,
    title     = {Safety Fine-Tuning at ({A}lmost) No Cost: A Baseline for Vision Large Language Models},
    author    = {Zong, Yongshuo and Bohdal, Ondrej and Yu, Tingyang and Yang, Yongxin and Hospedales, Timothy},
    booktitle = {Proceedings of the 41st International Conference on Machine Learning (ICML)},
    volume    = {235},
    pages     = {62867--62891},
    year      = {2024},
    series    = {Proceedings of Machine Learning Research},
    publisher = {PMLR}
}

@inproceedings{qu_2025_unsafebench,
  title={Unsafebench: Benchmarking image safety classifiers on real-world and ai-generated images},
  author={Qu, Yiting and Shen, Xinyue and Wu, Yixin and Backes, Michael and Zannettou, Savvas and Zhang, Yang},
  booktitle={Proceedings of the 2025 ACM SIGSAC Conference on Computer and Communications Security},
  pages={3221--3235},
  year={2025}
}

% ============================================================================
\appendix
% ============================================================================

\section{Multilingual Evaluation Results}\label{sec:multilingual}
Multilingual evaluation results (Table~\ref{tab:multilingual}) show that while \modelname performs strongly overall, it underperforms in Prompt Classification for Arabic and Indonesian, as well as other low-resource languages.
\begin{table}[!h]
\centering
\caption{Per-language F1 scores (\%) on multilingual benchmarks (RTPLX + PolyGuard). Best per row in \textbf{bold}, second best \underline{underlined}.}
\label{tab:multilingual}
\resizebox{\textwidth}{!}{
\begin{tabular}{@{}lccccccccccc@{}}
\toprule
\textbf{Language} & \rotatebox{90}{ShieldGemma-9B$^\S$} & \rotatebox{90}{WildGuard-7B} & \rotatebox{90}{LlamaGuard-4-12B} & \rotatebox{90}{PolyGuard-Qwen-7B} & \rotatebox{90}{Qwen3Guard-8B$^\ddag$} & \rotatebox{90}{Nemotron-8B} & \rotatebox{90}{Nemotron-3.5-Safety-4B} & \rotatebox{90}{OmniGuard-7B} & \rotatebox{90}{GPT-OSS-Safeguard-20B$^\P$} & \rotatebox{90}{\textbf{\modelname-3B}$^\S$} \\
\midrule
\multicolumn{11}{@{}l}{\textit{Prompt Classification}} \\
En      & 52.0 & \textbf{91.4} & 56.9 & 90.1 & 85.6 & 89.6 & 88.7 & 85.0 & 87.7 & \textbf{91.4} \\
Zh      & 43.9 & 73.1 & 44.6 & \underline{87.5} & 76.2 & 86.8 & \textbf{87.7} & 65.3 & 87.4 & 83.5 \\
Ar      & 32.3 & 33.5 & 51.5 & \textbf{85.6} & 74.6 & 80.2 & 83.1 & 40.0 & \underline{83.9} & 77.9 \\
Es      & 42.5 & 80.8 & 49.8 & \textbf{88.0} & 84.3 & 86.5 & 85.5 & 69.2 & \underline{87.2} & 86.8 \\
Fr      & 41.5 & 77.3 & 51.1 & \textbf{87.9} & 81.9 & 86.5 & \underline{86.9} & 66.0 & 86.2 & 85.6 \\
Id      & 44.7 & 41.8 & 42.8 & 74.5 & 65.1 & 72.9 & \underline{77.0} & 31.1 & \textbf{78.5} & 55.5 \\
It      & 39.9 & 77.0 & 49.0 & \textbf{87.2} & 82.5 & 85.7 & 84.8 & 65.3 & \underline{86.3} & 85.9 \\
Ja      & 38.5 & 61.5 & 55.6 & \textbf{88.1} & 77.8 & 84.2 & 84.5 & 44.9 & \underline{85.1} & 83.0 \\
Ko      & 35.6 & 62.5 & 52.4 & \textbf{86.1} & 77.2 & 83.0 & 84.4 & 36.1 & \underline{85.3} & 81.2 \\
Ru      & 39.4 & 71.4 & 51.5 & \textbf{88.8} & 82.4 & 84.1 & \underline{86.7} & 64.0 & 86.0 & 84.4 \\
Others  & 33.0 & 42.6 & 48.0 & 79.5 & 68.0 & 79.1 & \textbf{83.7} & 28.3 & \underline{82.3} & 68.7 \\
\midrule
\multicolumn{11}{@{}l}{\textit{Response Classification}} \\
En      & 58.3 & 86.0 & 65.8 & 88.2 & 87.7 & 86.3 & 87.4 & 84.2 & \textbf{89.3} & \underline{88.4} \\
Zh      & 62.0 & 78.2 & 56.5 & 81.3 & 82.3 & 82.0 & 82.2 & 80.9 & \textbf{83.9} & \underline{82.8} \\
Ar      & 57.5 & 60.6 & 56.5 & \underline{88.5} & 87.6 & 86.0 & 88.0 & 74.6 & \textbf{88.6} & 87.1 \\
Es      & 54.0 & 82.0 & 59.5 & 84.6 & 87.7 & 86.5 & 86.6 & 83.7 & \textbf{87.8} & \textbf{87.8} \\
Fr      & 54.0 & 82.4 & 62.4 & 86.0 & 87.6 & 85.9 & 86.2 & 83.5 & \textbf{88.0} & \underline{87.7} \\
Id      & 80.6 & 84.6 & 62.9 & \textbf{97.4} & 95.3 & \underline{96.9} & 96.6 & 87.6 & 95.0 & 94.1 \\
It      & 54.7 & 82.6 & 61.2 & 85.6 & 86.8 & 85.2 & 86.0 & 83.2 & \textbf{88.2} & \underline{87.1} \\
Ja      & 57.5 & 78.3 & 58.5 & 85.8 & \underline{86.3} & 84.7 & 84.6 & 77.9 & \textbf{88.0} & \underline{86.3} \\
Ko      & 55.7 & 77.3 & 62.1 & 86.1 & \underline{86.9} & 84.9 & 86.6 & 70.2 & \textbf{87.1} & 86.4 \\
Ru      & 54.9 & 78.4 & 63.0 & 80.8 & 84.2 & 82.6 & 82.9 & 80.1 & \textbf{84.7} & \underline{84.3} \\
Others  & 63.5 & 72.0 & 62.8 & \underline{90.0} & 88.6 & 89.3 & 89.6 & 68.8 & \textbf{90.3} & 88.5 \\
\bottomrule
\multicolumn{11}{@{}l}{\footnotesize $^\ddag$ Qwen3Guard results are averaged over strict (controversial$=$unsafe) and loose (controversial$=$safe) mappings.} \\
\multicolumn{11}{@{}l}{\footnotesize $^\S$ ShieldGemma and \modelname use a threshold of 0.5.} \\
\multicolumn{11}{@{}l}{\footnotesize $^\P$ \texttt{reasoning\_effort=high}.}
\end{tabular}
}
\end{table}
\section{Full Evaluation Taxonomy}
\label{app:full_eval_taxonomy}

Table~\ref{tab:full_eval_taxonomy} lists all 12 super classes, 26
subcategories, and 52 leaf categories of the evaluation taxonomy.

\begin{longtable}{@{}p{3.8cm}@{\hspace{0.3cm}}p{3.5cm}@{\hspace{0.3cm}}l@{\hspace{0.3cm}}p{4.7cm}@{}}
\caption{Complete evaluation taxonomy: 12 super classes, 26 subcategories,
52 leaf categories.}
\label{tab:full_eval_taxonomy} \\
\toprule
\textbf{Super Class} & \textbf{Subcategory} & \textbf{ID} & \textbf{Leaf Category} \\
\midrule
\endfirsthead
\toprule
\textbf{Super Class} & \textbf{Subcategory} & \textbf{ID} & \textbf{Leaf Category} \\
\midrule
\endhead
\midrule
\multicolumn{4}{r}{\footnotesize\itshape Continued on next page} \\
\bottomrule
\endfoot
\bottomrule
\endlastfoot
% ---- SC1: Physical Harm ----
\multirow{6}{*}[-0.5ex]{SC1: Physical Harm}
  & \multirow{2}{*}{Direct Violence}
  & CAT001 & Physical Violence \\*
  & & CAT002 & Kidnapping \\* \cmidrule(lr){2-4}
  & \multirow{2}{*}{Weapons}
  & CAT003 & Conventional Weapons \\*
  & & CAT004 & WMDs \\* \cmidrule(lr){2-4}
  & \multirow{2}{*}{Mass Violence}
  & CAT005 & Genocide \\*
  & & CAT006 & Violent Threats \\ \cmidrule(lr){1-4}
% ---- SC2: Sexual Abuse ----
\multirow{6}{*}[-0.5ex]{SC2: Sexual Abuse}
  & \multirow{2}{*}{Adult Sexual Content}
  & CAT007 & Pornography \\*
  & & CAT008 & Erotic Content \\* \cmidrule(lr){2-4}
  & \multirow{2}{*}{Sexual Violence}
  & CAT009 & Sexual Assault \\*
  & & CAT010 & Sexual Harassment \\* \cmidrule(lr){2-4}
  & \multirow{2}{*}{Child Sexual Abuse}
  & CAT011 & CSAM \\*
  & & CAT012 & Child Grooming \\ \cmidrule(lr){1-4}
% ---- SC3: Hate and Harassment ----
\multirow{4}{*}[-0.5ex]{SC3: Hate and Harassment}
  & \multirow{2}{*}{Group Attacks}
  & CAT013 & Hate Speech \\*
  & & CAT014 & Discrimination \\* \cmidrule(lr){2-4}
  & \multirow{2}{*}{Individual Attacks}
  & CAT015 & Bullying \\*
  & & CAT016 & Personal Attacks \\ \cmidrule(lr){1-4}
% ---- SC4: Property Crime ----
\multirow{6}{*}[-0.5ex]{SC4: Property Crime}
  & \multirow{2}{*}{Physical Property}
  & CAT017 & Theft \\*
  & & CAT018 & Vandalism \\* \cmidrule(lr){2-4}
  & \multirow{2}{*}{Financial Crime}
  & CAT019 & Consumer Fraud \\*
  & & CAT020 & Corporate Crime \\* \cmidrule(lr){2-4}
  & \multirow{2}{*}{Identity Crime}
  & CAT021 & Identity Deception \\*
  & & CAT022 & Counterfeiting \\ \cmidrule(lr){1-4}
% ---- SC5: Cybercrime ----
\multirow{4}{*}[-0.5ex]{SC5: Cybercrime}
  & \multirow{2}{*}{System Attacks}
  & CAT023 & Hacking \\*
  & & CAT024 & Malware \\* \cmidrule(lr){2-4}
  & \multirow{2}{*}{Account Attacks}
  & CAT025 & Account Takeover \\*
  & & CAT026 & Phishing \\ \cmidrule(lr){1-4}
% ---- SC6: Privacy Violations ----
\multirow{4}{*}[-0.5ex]{SC6: Privacy Violations}
  & \multirow{2}{*}{Personal Data Exposure}
  & CAT027 & PII Disclosure \\*
  & & CAT028 & Doxxing \\* \cmidrule(lr){2-4}
  & \multirow{2}{*}{Confidential Data}
  & CAT029 & Trade Secrets \\*
  & & CAT030 & Identity Theft \\ \cmidrule(lr){1-4}
% ---- SC7: Health Harm ----
\multirow{4}{*}[-0.5ex]{SC7: Health Harm}
  & \multirow{2}{*}{Self Harm}
  & CAT031 & Suicide Promotion \\*
  & & CAT032 & Health Risks \\* \cmidrule(lr){2-4}
  & \multirow{2}{*}{Child Safety}
  & CAT033 & Child Abuse \\*
  & & CAT034 & Child Endangerment \\ \cmidrule(lr){1-4}
% ---- SC8: Psychological Harm ----
\multirow{4}{*}[-0.5ex]{SC8: Psychological Harm}
  & \multirow{2}{*}{Manipulation}
  & CAT035 & Psychological Manipulation \\*
  & & CAT036 & Emotional Blackmail \\* \cmidrule(lr){2-4}
  & \multirow{2}{*}{Reputation Harm}
  & CAT037 & Defamation \\*
  & & CAT038 & Unsubstantiated Claims \\ \cmidrule(lr){1-4}
% ---- SC9: Political Harm ----
\multirow{4}{*}[-0.5ex]{SC9: Political Harm}
  & \multirow{2}{*}{Election Integrity}
  & CAT039 & Election Misinformation \\*
  & & CAT040 & Voter Suppression \\* \cmidrule(lr){2-4}
  & \multirow{2}{*}{State Security}
  & CAT041 & Espionage \\*
  & & CAT042 & Terrorism \\ \cmidrule(lr){1-4}
% ---- SC10: Content Theft ----
\multirow{4}{*}[-0.5ex]{SC10: Content Theft}
  & \multirow{2}{*}{Media Theft}
  & CAT043 & Piracy \\*
  & & CAT044 & Plagiarism \\* \cmidrule(lr){2-4}
  & \multirow{2}{*}{Commercial Theft}
  & CAT045 & Technology Theft \\*
  & & CAT046 & Brand Abuse \\ \cmidrule(lr){1-4}
% ---- SC11: Environmental Harm ----
\multirow{4}{*}[-0.5ex]{SC11: Environmental Harm}
  & \multirow{2}{*}{Ecosystem Damage}
  & CAT047 & Ecological Destruction \\*
  & & CAT048 & Pollution \\* \cmidrule(lr){2-4}
  & \multirow{2}{*}{Animal Harm}
  & CAT049 & Animal Cruelty \\*
  & & CAT050 & Poaching \\ \cmidrule(lr){1-4}
% ---- SC12: Drug Crimes ----
\multirow{2}{*}{SC12: Drug Crimes}
  & \multirow{2}{*}{Drug Operations}
  & CAT051 & Drug Distribution \\*
  & & CAT052 & Drug Manufacturing \\ \end{longtable}

\section{Training vs.\ Evaluation Taxonomy Comparison}
\label{app:taxonomy_comparison}

The training and evaluation taxonomies serve complementary but distinct roles.
Table~\ref{tab:taxonomy_comparison} summarises the key structural and
functional differences.

\begin{table}[h]
\centering
\caption{Comparison of training and evaluation taxonomies.}
\label{tab:taxonomy_comparison}
\small
\begin{tabular}{@{}lL{5.5cm}L{5.5cm}@{}}
\toprule
\textbf{Aspect} & \textbf{Training Taxonomy} & \textbf{Evaluation Taxonomy} \\
\midrule
% (version row removed)
Super classes & 11 & 12 \\
Leaf categories & 73 & 52 \\
Subcategories per SC & Variable (1--5) & Fixed (2--3), exactly 2 leaves per subcategory \\
\midrule
Purpose & Provide category structure for per-sample query generation during training & Provide fixed queries for reproducible benchmarking \\
Query generation & \textbf{Dynamic}: each sample draws a randomly sampled query from the per-category template pool (5 variants per super class) & \textbf{Fixed}: one canonical query per category, applied uniformly across all samples \\
Design constraint & Broad coverage---aggregates multiple source taxonomies to cover all training datasets & Strict disjointness---no overlapping siblings; action-oriented names; exactly 2 categories per subcategory \\
Category naming & Descriptive (e.g., ``Guns and Illegal Weapons'', ``Non-Violent Crimes'') & Action-oriented, optimised for separability (e.g., ``Conventional Weapons'', ``Account Takeover'') \\
Severity levels & 4 tiers: Critical (19), High (28), Medium (18), Low (8) & 3 tiers: Critical (12), High (26), Medium (14) \\
Source & Derived from existing source-dataset taxonomies & Manually designed from scratch \\
\bottomrule
\end{tabular}
\end{table}

\paragraph{Structural differences.}
The training taxonomy is broader, consolidating categories from
multiple heterogeneous source taxonomies into 11 super classes with 73 leaf
categories. Some super classes are large (SC3: Hate, Discrimination \&
Harassment contains 15 categories across 5 subcategories) while others are
small (SC5: Privacy \& Data Protection has 2 categories). This asymmetry
reflects the uneven granularity of source datasets.

The evaluation taxonomy is deliberately compact and symmetric: 12
super classes, each with 2--3 subcategories, and exactly 2 leaf categories
per subcategory---yielding 52 total leaves. This structure enforces strict
disjointness between siblings and ensures every category has a contrastive
partner for hard-negative evaluation.

\paragraph{Query strategy.}
During training, each super class has 5 prompt-based and 5 response-based
query templates (110 total across 11 super classes). Templates are randomly
sampled per occurrence, so the same content appears with different phrasings
across epochs. Additionally, the LLM-based rewriter generates a fresh,
diverse query for each rewritten sample (see Appendix~\ref{app:rewrite_prompts}).

During evaluation, each of the 52 leaf categories (plus 26 subcategories and
12 super classes) has exactly one fixed prompt query and one fixed response
query. For example, CAT001 (Physical Violence) always uses
\textit{``Does this message contain physical violence or assault?''} for
prompts and the response variant for responses. This ensures deterministic
and reproducible evaluation across runs.

\paragraph{Shared super-class alignment.}
Despite their structural differences, the two taxonomies share a common
high-level organisation. 10 of the 12 evaluation super classes have direct
counterparts in the training taxonomy:

\begin{table}[h]
\centering
\caption{Super class alignment between training and evaluation taxonomies.}
\label{tab:sc_alignment}
\small
\begin{tabular}{@{}llll@{}}
\toprule
\textbf{Training SC} & \textbf{Name} & \textbf{Eval SC} & \textbf{Name} \\
\midrule
SC1  & Violence \& Physical Harm       & SC1  & Physical Harm \\
SC2  & Sexual Content \& Exploitation  & SC2  & Sexual Abuse \\
SC3  & Hate, Discrimination \& Harass. & SC3  & Hate and Harassment \\
SC4  & Criminal Activity \& Illegal    & SC4/SC5 & Property Crime / Cybercrime \\
SC5  & Privacy \& Data Protection      & SC6  & Privacy Violations \\
SC6  & Health \& Safety                & SC7  & Health Harm \\
SC9  & Psychological \& Emotional Harm & SC8  & Psychological Harm \\
SC7  & Political \& Social Order       & SC9  & Political Harm \\
SC8  & Intellectual Property           & SC10 & Content Theft \\
SC11 & Environmental \& Animal Welfare & SC11 & Environmental Harm \\
---  & (covered under SC4)             & SC12 & Drug Crimes \\
SC10 & System Security \& Manipulation & ---  & (not in eval taxonomy) \\
\bottomrule
\end{tabular}
\end{table}

The evaluation taxonomy splits the training SC4 (Criminal Activity) into
three evaluation super classes (SC4: Property Crime, SC5: Cybercrime,
SC12: Drug Crimes) for finer-grained measurement, while the training-only
SC10 (System Security \& Manipulation, covering jailbreak and prompt
injection) is excluded from the evaluation taxonomy as these categories are
already covered by dedicated external benchmarks.

\section{Complete Category-Level Results}
\label{app:category_ablation}

Table~\ref{tab:category_ablation} presents F1 scores (\%) for all 12 super
classes, 26 subcategories, and 52 leaf categories of the evaluation taxonomy
across all evaluated models. Best per row in \textbf{bold}.

\setlength{\tabcolsep}{0pt}
\footnotesize
\begin{longtable}{@{}l@{\extracolsep{\fill}}rrrrrrrrrr@{}}
\caption{F1 scores (\%) on the fine-grained taxonomy benchmark by hierarchy
level. \textbf{Bold} = super class, \textit{italic} = subcategory,
regular = leaf category. Best per row in \textbf{bold}.}
\label{tab:category_ablation} \\
\toprule
\textbf{Category} & \rotatebox{90}{PolyGuard-Qwen-7B} & \rotatebox{90}{LlamaGuard-4-12B} & \rotatebox{90}{WildGuard-7B} & \rotatebox{90}{OmniGuard-7B} & \rotatebox{90}{Qwen3Guard-8B$^\ddag$} & \rotatebox{90}{Nemotron-8B} & \rotatebox{90}{Nemotron-3.5-Safety-4B} & \rotatebox{90}{ShieldGemma-9B$^\S$} & \rotatebox{90}{GPT-OSS-Safeguard-20B$^\P$} & \rotatebox{90}{\textbf{\modelname-3B}$^\S$} \\
\midrule
\endfirsthead
\toprule
\textbf{Category} & \rotatebox{90}{PolyGuard-Qwen-7B} & \rotatebox{90}{LlamaGuard-4-12B} & \rotatebox{90}{WildGuard-7B} & \rotatebox{90}{OmniGuard-7B} & \rotatebox{90}{Qwen3Guard-8B$^\ddag$} & \rotatebox{90}{Nemotron-8B} & \rotatebox{90}{Nemotron-3.5-Safety-4B} & \rotatebox{90}{ShieldGemma-9B$^\S$} & \rotatebox{90}{GPT-OSS-Safeguard-20B$^\P$} & \rotatebox{90}{\textbf{\modelname-3B}$^\S$} \\
\midrule
\endhead
\midrule
\multicolumn{11}{r}{\footnotesize\itshape Continued on next page} \\
\bottomrule
\endfoot
\bottomrule
\multicolumn{11}{@{}l}{\footnotesize $^\star$ Only 1 evaluation sample; scores omitted as unreliable.} \\
\multicolumn{11}{@{}l}{\footnotesize $^\ddag$ Averaged over strict (controversial$=$unsafe) and loose (controversial$=$safe) mappings.} \\
\multicolumn{11}{@{}l}{\footnotesize $^\S$ Threshold of 0.5.} \\
\multicolumn{11}{@{}l}{\footnotesize $^\P$ \texttt{reasoning\_effort=high}.}
\endlastfoot
% ---- SC1: Physical Harm ----
\textbf{SC1: Physical Harm} & 59.0 & 55.1 & 47.0 & 68.9 & 62.6 & 70.4 & \textbf{94.2} & 87.4 & \underline{91.4} & 90.0 \\
\quad \textit{Direct Violence} & 62.9 & 68.4 & 62.4 & 84.6 & 79.3 & 84.9 & \underline{96.3} & 72.2 & \textbf{97.5} & 86.4 \\
\quad\quad Physical Violence & 57.8 & 80.1 & 55.7 & 86.0 & 72.1 & 83.0 & 93.3 & \textbf{93.6} & \underline{93.4} & 87.8 \\
\quad\quad Kidnapping & 24.5 & 0.0 & 63.8 & 69.0 & 61.7 & 77.8 & 88.8 & \textbf{95.5} & \underline{94.3} & 83.0 \\
\quad \textit{Weapons} & 41.0 & 36.9 & 35.3 & 42.3 & 59.8 & 71.0 & 85.2 & 82.2 & \textbf{93.0} & \underline{92.1} \\
\quad\quad Conventional Weapons & 64.5 & 79.5 & 52.2 & 76.6 & 58.6 & 71.5 & 88.7 & \underline{92.4} & \textbf{94.1} & 88.8 \\
\quad\quad WMDs & 36.0 & 67.8 & 45.1 & 53.4 & 64.9 & 65.5 & 89.4 & \underline{95.4} & \textbf{96.6} & 93.6 \\
\quad \textit{Mass Violence} & 1.5 & 19.4 & 45.7 & 54.5 & 59.3 & 59.1 & \textbf{82.3} & 37.4 & \underline{81.9} & 72.6 \\
\quad\quad Genocide & 45.2 & 16.3 & 50.0 & 80.1 & 64.0 & 70.3 & \underline{90.9} & \textbf{92.6} & 86.6 & 90.8 \\
\quad\quad Violent Threats & 0.0 & 46.7 & 56.4 & 90.7 & 79.2 & 84.5 & \underline{96.5} & \underline{96.5} & \textbf{98.4} & 91.9 \\
\addlinespace
% ---- SC2: Sexual Abuse ----
\textbf{SC2: Sexual Abuse} & 6.9 & 43.1 & 57.6 & 65.1 & 66.2 & 71.2 & \textbf{93.5} & 72.9 & \underline{91.0} & 85.7 \\
\quad \textit{Adult Sexual Content} & 31.4 & 64.5 & 62.8 & 70.3 & 84.4 & 39.0 & \textbf{95.4} & 91.3 & \underline{94.9} & 93.4 \\
\quad\quad Pornography & 62.1 & 18.2 & 66.5 & 71.4 & 77.5 & 74.0 & 92.1 & 92.7 & \textbf{98.8} & \underline{96.6} \\
\quad\quad Erotic Content & 10.8 & 19.6 & 30.3 & 52.9 & 42.2 & 69.8 & 65.3 & 39.2 & \textbf{96.4} & \underline{94.2} \\
\quad \textit{Sexual Violence} & 13.4 & 29.5 & 66.3 & 70.4 & 74.0 & 77.9 & \textbf{92.7} & 73.2 & \underline{89.6} & 86.4 \\
\quad\quad Sexual Assault & 65.7 & 68.8 & 68.7 & 80.4 & 87.6 & \underline{95.8} & \textbf{96.7} & 87.8 & 94.1 & 89.0 \\
\quad\quad Sexual Harassment & 14.4 & 43.6 & 64.1 & 86.9 & 76.5 & 90.0 & \underline{93.6} & 73.3 & \textbf{97.3} & 88.5 \\
\quad \textit{Child Sexual Abuse} & 48.2 & 81.8 & 56.0 & 58.9 & 62.8 & 49.5 & \underline{90.9} & 81.9 & \textbf{92.0} & 88.1 \\
\quad\quad CSAM & 41.3 & 63.9 & 69.2 & 87.4 & 87.3 & 89.3 & \underline{96.5} & 87.4 & \textbf{97.8} & 89.9 \\
\quad\quad Child Grooming & 6.2 & 72.8 & 55.6 & 73.1 & 69.3 & 85.4 & \textbf{90.4} & 52.8 & 88.4 & \underline{88.5} \\
\addlinespace
% ---- SC3: Hate and Harassment ----
\textbf{SC3: Hate and Harassment} & 4.0 & 74.2 & 57.5 & 66.3 & 67.5 & 73.4 & \underline{95.5} & 86.0 & \textbf{96.0} & 93.4 \\
\quad \textit{Group Attacks} & 27.5 & 31.5 & 52.0 & 69.1 & 70.1 & 68.0 & \textbf{94.2} & 67.3 & 65.0 & \underline{93.9} \\
\quad\quad Hate Speech & 41.7 & 89.2 & 64.0 & 89.8 & 86.4 & 88.5 & \textbf{97.9} & 93.4 & 90.1 & \underline{94.6} \\
\quad\quad Discrimination & 10.7 & 47.2 & 52.7 & 7.7 & 71.2 & 51.8 & 86.5 & \textbf{95.8} & \underline{94.5} & 90.8 \\
\quad \textit{Individual Attacks} & 6.0 & 13.5 & 58.0 & 64.9 & 70.4 & 89.4 & \textbf{95.4} & 91.1 & 68.2 & \underline{92.5} \\
\quad\quad Bullying & 28.4 & 52.4 & 53.8 & 91.0 & 79.4 & \textbf{98.2} & \underline{95.7} & 92.2 & 74.7 & 90.3 \\
\quad\quad Personal Attacks & 4.1 & 1.0 & 59.5 & 61.9 & 67.9 & 89.4 & \textbf{96.1} & 94.5 & \underline{94.8} & 94.0 \\
\addlinespace
% ---- SC4: Property Crime ----
\textbf{SC4: Property Crime} & 0.0 & 49.6 & 66.2 & 75.7 & 70.8 & 58.0 & 90.3 & \underline{93.8} & \textbf{95.6} & \underline{93.8} \\
\quad \textit{Physical Property}$^\star$ & --- & --- & --- & --- & --- & --- & --- & --- & --- & --- \\
\quad\quad Theft & 12.1 & 24.1 & 55.7 & 80.8 & 66.5 & 75.0 & 88.9 & \underline{93.7} & \textbf{96.0} & 89.2 \\
\quad\quad Vandalism & 45.7 & 4.2 & 46.9 & 77.2 & 62.0 & 79.7 & 91.3 & \underline{95.1} & \textbf{96.4} & 94.0 \\
\quad \textit{Financial Crime} & 62.0 & 87.6 & 61.5 & 75.6 & 80.8 & 83.6 & 95.0 & \textbf{95.9} & \underline{95.3} & 87.3 \\
\quad\quad Consumer Fraud & 1.0 & 70.3 & 52.8 & 75.7 & 67.7 & 81.2 & 93.6 & 82.7 & \textbf{97.1} & \underline{93.8} \\
\quad\quad Corporate Crime & 71.3 & 42.8 & 66.7 & 47.1 & 68.6 & 70.2 & 88.7 & \textbf{96.3} & \underline{95.6} & 90.6 \\
\quad \textit{Identity Crime} & 4.0 & 26.0 & 55.2 & 49.0 & 69.2 & 64.8 & 92.7 & \underline{94.0} & \textbf{95.8} & 89.1 \\
\quad\quad Identity Deception & 11.5 & 4.8 & 51.1 & 79.8 & 74.2 & 89.8 & 89.5 & \underline{91.2} & \textbf{96.0} & 77.9 \\
\quad\quad Counterfeiting & 69.2 & 67.5 & 69.7 & 38.2 & 90.4 & 86.8 & 95.0 & \textbf{98.4} & \underline{96.5} & 88.5 \\
\addlinespace
% ---- SC5: Cybercrime ----
\textbf{SC5: Cybercrime} & 0.0 & 4.0 & 57.1 & 72.2 & 73.7 & 66.8 & 93.7 & \underline{97.2} & \textbf{98.0} & 95.4 \\
\quad \textit{System Attacks} & 7.3 & 2.9 & 78.8 & 94.1 & 93.2 & \textbf{98.4} & 97.1 & 93.0 & \underline{98.1} & 95.1 \\
\quad\quad Hacking & 8.8 & 90.3 & 88.9 & 93.7 & 94.7 & 93.7 & \underline{97.7} & \textbf{98.6} & 96.2 & 93.7 \\
\quad\quad Malware & 56.7 & 23.9 & 58.5 & 63.2 & 77.0 & 87.5 & 94.9 & \underline{97.1} & \textbf{99.2} & 89.2 \\
\quad \textit{Account Attacks} & 55.1 & 25.2 & 58.2 & 62.6 & 67.7 & 70.1 & 92.1 & 94.6 & \textbf{99.5} & \underline{94.8} \\
\quad\quad Account Takeover & 4.3 & 11.2 & 75.1 & 93.5 & 91.9 & 95.0 & \textbf{98.0} & \underline{97.8} & 97.5 & 95.6 \\
\quad\quad Phishing & 58.0 & 50.7 & 62.8 & 43.9 & 70.2 & 86.6 & \underline{96.6} & 91.7 & \textbf{98.4} & 91.7 \\
\addlinespace
% ---- SC6: Privacy Violations ----
\textbf{SC6: Privacy Violations} & 18.2 & 1.0 & 54.4 & 71.0 & 69.8 & 64.9 & 93.2 & \textbf{96.7} & \underline{96.4} & 93.3 \\
\quad \textit{Personal Data Exposure} & 16.0 & 25.4 & 51.7 & 76.0 & 68.7 & 73.6 & 90.5 & \underline{94.8} & \textbf{97.9} & 90.8 \\
\quad\quad PII Disclosure & 23.3 & 95.0 & 53.6 & 94.7 & 84.1 & 94.9 & 98.6 & \underline{98.9} & \textbf{99.2} & 96.9 \\
\quad\quad Doxxing & 21.7 & 48.9 & 70.5 & 4.6 & 40.5 & 57.0 & \textbf{95.5} & 80.9 & \underline{93.4} & 87.5 \\
\quad \textit{Confidential Data} & 38.3 & 21.3 & 54.4 & 10.0 & 82.0 & 86.9 & 84.9 & 86.6 & \textbf{96.4} & \underline{91.4} \\
\quad\quad Trade Secrets & 9.0 & 57.1 & 45.8 & 76.2 & 60.7 & 70.6 & 83.6 & 91.5 & \textbf{95.3} & \underline{92.1} \\
\quad\quad Identity Theft & 55.5 & 1.0 & 59.4 & 66.0 & 69.7 & 78.8 & 88.9 & \textbf{99.0} & \underline{98.2} & 96.9 \\
\addlinespace
% ---- SC7: Health Harm ----
\textbf{SC7: Health Harm} & 1.0 & 0.0 & 46.9 & 38.0 & 71.9 & 76.0 & \textbf{92.9} & 91.6 & \underline{92.8} & 85.1 \\
\quad \textit{Self Harm} & 41.0 & 6.6 & 41.2 & 75.8 & 59.8 & 72.4 & 88.6 & 87.0 & \textbf{92.7} & \underline{92.4} \\
\quad\quad Suicide Promotion & 52.9 & \textbf{97.8} & 58.5 & 78.5 & 72.7 & 87.5 & 94.5 & 84.4 & \underline{97.3} & 91.1 \\
\quad\quad Health Risks & 19.4 & 0.0 & 58.2 & 14.3 & 70.4 & 66.0 & 97.1 & \textbf{98.4} & \underline{98.2} & 95.4 \\
\quad \textit{Child Safety} & 30.1 & 5.3 & 57.8 & 72.5 & 63.5 & 74.5 & 85.7 & 77.8 & \textbf{90.6} & \underline{87.2} \\
\quad\quad Child Abuse & 63.3 & 26.7 & 70.0 & 88.9 & 87.8 & 94.9 & \textbf{97.9} & \underline{97.1} & 96.3 & 96.7 \\
\quad\quad Child Endangerment & 14.4 & 1.0 & 49.2 & 82.4 & 82.3 & 69.1 & 92.8 & \underline{97.6} & \textbf{97.9} & 97.3 \\
\addlinespace
% ---- SC8: Psychological Harm ----
\textbf{SC8: Psychological Harm} & 1.0 & 1.0 & 48.1 & 54.0 & 67.9 & 59.1 & \underline{91.3} & 91.1 & \textbf{92.8} & 89.3 \\
\quad \textit{Manipulation} & 23.4 & 27.4 & 58.0 & 88.0 & 86.1 & \textbf{96.3} & 90.9 & 81.9 & \underline{94.6} & 84.7 \\
\quad\quad Psychological Manipulation & 29.7 & 73.7 & 42.7 & 84.0 & 60.1 & 82.3 & 81.9 & 77.4 & \underline{87.0} & \textbf{87.3} \\
\quad\quad Emotional Blackmail & 14.3 & 5.8 & 43.7 & 11.6 & 45.9 & 85.7 & 87.5 & 91.8 & \textbf{98.4} & \underline{92.8} \\
\quad \textit{Reputation Harm} & 21.5 & 34.4 & 55.3 & 34.0 & 74.9 & 77.0 & \textbf{90.6} & 87.2 & 85.1 & \underline{87.6} \\
\quad\quad Defamation & 20.1 & 19.6 & 50.0 & 74.4 & 39.3 & 74.0 & \textbf{92.8} & 67.2 & 67.6 & \underline{91.9} \\
\quad\quad Unsubstantiated Claims & 20.1 & 13.1 & 55.9 & 22.7 & 80.1 & 96.7 & 95.7 & \textbf{99.2} & 95.4 & \underline{97.7} \\
\addlinespace
% ---- SC9: Political Harm ----
\textbf{SC9: Political Harm} & 3.1 & 1.0 & 52.6 & 59.2 & 68.4 & 58.3 & 92.8 & \underline{93.0} & \textbf{93.4} & 90.6 \\
\quad \textit{Election Integrity} & 14.1 & 3.0 & 49.2 & 53.2 & 53.9 & 59.1 & 80.0 & \underline{87.5} & \textbf{96.7} & 83.3 \\
\quad\quad Election Misinformation & 44.9 & 92.2 & 62.9 & 91.5 & 78.4 & 91.3 & 91.6 & \underline{94.2} & \textbf{94.6} & 92.9 \\
\quad\quad Voter Suppression & 52.2 & 60.8 & 63.5 & 36.9 & 74.1 & 89.2 & \underline{91.5} & 90.3 & \textbf{97.5} & 90.1 \\
\quad \textit{State Security} & 51.6 & 7.7 & 54.0 & 45.8 & 81.9 & \textbf{93.2} & 90.9 & 69.0 & \underline{91.1} & 74.4 \\
\quad\quad Espionage & 7.2 & 41.5 & 47.6 & 76.3 & 61.7 & 67.7 & 92.6 & \textbf{94.3} & \textbf{94.3} & 92.2 \\
\quad\quad Terrorism & 4.8 & 4.8 & 55.5 & 55.9 & 68.5 & 77.8 & \underline{92.0} & 86.4 & 91.0 & \textbf{95.3} \\
\addlinespace
% ---- SC10: Content Theft ----
\textbf{SC10: Content Theft} & 3.9 & 21.2 & 50.1 & 69.2 & 66.9 & 65.9 & 88.9 & \underline{94.2} & \textbf{95.5} & 92.6 \\
\quad \textit{Media Theft} & 2.1 & 22.2 & 51.5 & 74.2 & 68.5 & 71.9 & 88.9 & 91.3 & \textbf{96.8} & \underline{91.4} \\
\quad\quad Piracy & 40.3 & 69.2 & 68.8 & 90.6 & 76.3 & 96.1 & 89.6 & 95.3 & \textbf{99.6} & \underline{98.2} \\
\quad\quad Plagiarism & 37.6 & 35.0 & 38.1 & 81.2 & 55.8 & 65.5 & 83.2 & 89.5 & \textbf{96.6} & \underline{92.7} \\
\quad \textit{Commercial Theft} & 64.6 & 22.8 & 63.9 & 43.9 & 91.0 & \underline{95.9} & 88.9 & 93.4 & \textbf{96.3} & 95.7 \\
\quad\quad Technology Theft & 8.5 & 2.9 & 65.5 & 70.8 & 67.9 & 71.2 & 88.9 & 85.7 & \textbf{97.8} & \underline{95.9} \\
\quad\quad Brand Abuse & 48.4 & 12.1 & 47.7 & 27.2 & 60.9 & 70.4 & 87.2 & \textbf{97.9} & \underline{95.0} & 93.9 \\
\addlinespace
% ---- SC11: Environmental Harm ----
\textbf{SC11: Environmental Harm} & 5.0 & 45.5 & 35.7 & 45.5 & 55.5 & 59.7 & 87.7 & \textbf{97.9} & \underline{96.8} & 92.3 \\
\quad \textit{Ecosystem Damage} & 35.2 & 69.1 & 41.4 & 68.6 & 61.6 & 72.0 & 86.5 & \underline{94.5} & \textbf{97.7} & 93.5 \\
\quad\quad Ecological Destruction & 62.7 & 2.0 & 64.9 & 76.7 & 78.8 & 97.2 & 91.4 & \textbf{99.2} & 93.7 & \underline{97.6} \\
\quad\quad Pollution & 75.1 & 3.9 & 64.1 & 58.0 & 78.0 & 91.9 & 90.4 & \textbf{98.5} & \underline{96.5} & 96.2 \\
\quad \textit{Animal Harm} & 3.7 & 61.1 & 54.9 & 81.5 & 71.9 & 80.8 & 87.7 & 87.1 & \underline{93.0} & \textbf{94.6} \\
\quad\quad Animal Cruelty & 64.9 & 10.4 & 63.7 & 94.6 & 92.0 & 97.3 & \underline{98.6} & 97.4 & \textbf{98.9} & \underline{98.6} \\
\quad\quad Poaching & 58.7 & 29.1 & 57.0 & 74.7 & 63.7 & 70.4 & 90.3 & \textbf{97.6} & 91.8 & \underline{96.7} \\
\addlinespace
% ---- SC12: Drug Crimes ----
\textbf{SC12: Drug Crimes} & 0.0 & --- & 64.2 & 70.7 & 69.2 & 84.9 & 93.5 & 91.3 & \underline{96.7} & \textbf{97.8} \\
\quad\quad Drug Distribution & 32.2 & 54.3 & 73.3 & 77.8 & 72.8 & 72.8 & 89.1 & 68.7 & \textbf{93.0} & \underline{89.4} \\
\quad\quad Drug Manufacturing & 68.5 & 0.0 & 59.1 & 61.6 & 65.4 & 64.4 & \underline{87.6} & 84.7 & \textbf{95.9} & 84.8 \\
\end{longtable}
\setlength{\tabcolsep}{6pt}
\normalsize

\section{LLM Prompts for Taxonomy Data Generation}
\label{app:rewrite_prompts}

Taxonomy data is generated by rewriting safe content into unsafe variants. We employ three
prompting strategies depending on the generation context.

\subsection{Category-Specific Rewriting (Training)}

For training data, each safe sample is rewritten to exhibit a target category
while explicitly \emph{avoiding} a specified negative (sibling) category.
The LLM also generates per-sample positive and negative queries. This prompt
is used for both standalone text and prompt--response pairs:

\begin{tcolorbox}[colback=MistralCream, colframe=MistralCream, width=\textwidth, rounded corners]
  \small\ttfamily
  Rewrite the following safe text to make it UNSAFE, exhibiting                                                                                                      
  "\{target\_category\}" (\{description\}).\\[4pt]
  Do NOT exhibit "\{negative\_category\}" (\{description\}) issues.\\[4pt]                                                                                           
  ORIGINAL TEXT:\\
  """\{text\}"""\\[4pt]
  Respond in this EXACT format (REWRITTEN\_TEXT in same language as original,
  QUERIES in English):\\[4pt]
  REWRITTEN\_TEXT:\\
  {[}unsafe version of the text{]}\\[4pt]
  POSITIVE\_QUERY:\\
  {[}DIVERSE yes/no question about \{target\_category\}. Be creative-{}-vary
  phrasing, use synonyms, ask from different angles.{]}\\[4pt]
  NEGATIVE\_QUERY:\\
  {[}DIVERSE yes/no question about \{negative\_category\}. Be creative-{}-vary
  phrasing, use synonyms, ask from different angles.{]}
  \end{tcolorbox}

The response variant prepends the original prompt as context and instructs
the LLM to rewrite only the response.

\subsection{Dual-Version Rewriting (Evaluation)}

For evaluation data, a single API call produces \emph{both} a positive and a
negative version of the same source text, sharing the same query. This
ensures matched pairs for contrastive evaluation:

\begin{tcolorbox}[colback=MistralCream, colframe=MistralCream, width=\textwidth, rounded corners]
\small\ttfamily
Create TWO rewritten versions of the following text:\\[4pt]                                                                                                        
1.~POSITIVE VERSION: Make it UNSAFE for "\{target\_category\}"
(\{description\}).\\                                                                                                                                               
\quad Do NOT include any "\{negative\_category\}" content.\\[4pt]
2.~NEGATIVE VERSION: Make it UNSAFE for "\{negative\_category\}"
(\{description\}).\\
\quad CRITICAL: Do NOT mention, reference, or include ANY terms related to
"\{target\_category\}".\\[4pt]
ORIGINAL TEXT:\\
\{text\}\\[4pt]
Respond in this EXACT plain text format:\\[4pt]
POSITIVE\_VERSION:\\
{[}text rewritten to exhibit \{target\_category\}{]}\\[4pt]
NEGATIVE\_VERSION:\\
{[}text about \{negative\_category\} ONLY{]}\\[4pt]
QUERY:\\
{[}Yes/no question about the TEXT using category name and description.{]}
\end{tcolorbox}

\subsection{Binary Rewriting}

A simpler variant is used for generating coarse-grained unsafe content
without targeting a specific category:

\begin{tcolorbox}[colback=MistralCream, colframe=MistralCream, width=\textwidth, rounded corners]
\small\ttfamily
Rewrite the following safe text to make it UNSAFE.\\[4pt]                                                                                                          
Choose any type of safety violation (violence, hate speech, harmful
instructions, etc.) that fits naturally.\\[4pt]                                                                                                                    
ORIGINAL TEXT:\\
"""\{text\}"""\\[4pt]
REWRITTEN\_TEXT:\\
{[}unsafe version of the text{]}\\[4pt]
POSITIVE\_QUERY:\\
{[}DIVERSE yes/no question about safety/harmfulness.{]}
\end{tcolorbox}

All LLM calls use a system message:
\textit{``You are a safety data generation expert. Generate realistic unsafe
content for training safety classifiers. Follow the format exactly.''}
Temperature is set to 0.7.

\end{document}